%% file: main.tex
\documentclass[journal]{IEEEtran}
\usepackage{cite}
\usepackage{float}
\usepackage{amsmath,amssymb,amsfonts}
\usepackage{algorithm}
\usepackage{algorithmic}
\usepackage{subfig}
\usepackage{array}
\usepackage{booktabs}
\usepackage{multirow}
\usepackage{graphicx}
\usepackage{textcomp}
\usepackage{xcolor}

\newtheorem{assumption}{Assumption}
\newtheorem{proposition}{Proposition}
\newtheorem{lemma}{Lemma}

\newtheorem{theorem}{Theorem}
\newtheorem{definition}{Definition}
\newtheorem{remark}{Remark}

\begin{document}

\title{On the Convergence Theory of Meta Reinforcement Learning with Personalized Policies}

% \author{Michael~Shell,~\IEEEmembership{Member,~IEEE,}
%         John~Doe,~\IEEEmembership{Fellow,~OSA,}
%         and~Jane~Doe,~\IEEEmembership{Life~Fellow,~IEEE}% <-this % stops a space
% \thanks{M. Shell was with the Department
% of Electrical and Computer Engineering, Georgia Institute of Technology, Atlanta,
% GA, 30332 USA e-mail: (see http://www.michaelshell.org/contact.html).}% <-this % stops a space
% \thanks{J. Doe and J. Doe are with Anonymous University.}% <-this % stops a space
% \thanks{Manuscript received April 19, 2005; revised August 26, 2015.}}

\author{Haozhi Wang, Qing Wang,~\IEEEmembership{Member,~IEEE},Yunfeng Shao, Dong Li, Jianye Hao, Yinchuan Li,~\IEEEmembership{Member,~IEEE}
    %<-this % stops a space
\thanks{\emph{Corresponding author: Yinchuan Li (email: liyinchuan@huawei.com).}}
\thanks{Haozhi Wang, Qing Wang and Jianye Hao are with School of Electrical and Information Engineering, Tianjin University, Tianjin, China. 
This work was completed while Haozhi Wang was a member of the Huawei Noah’s Ark Lab for Advanced Study.
}
\thanks{Yunfeng Shao, Dong Li and Yinchuan Li are with Huawei Noah's Ark Lab, Beijing, China}
        % <-this % stops a space
}

% This work was completed while Haozhi Wang was an intern of the Huawei Noah's Ark Lab.
% (e-mail: wanghaozhi@tju.edu.cn, wangq@tju.edu.cn, jianye.hao@tju.edu.cn)
% (e-mail: shaoyunfeng@huawei.com, lidong106@huawei.com, liyinchuan@huawei.com)

\markboth{Journal of \LaTeX\ Class Files,~Vol.~14, No.~8, August~2015}%
{Shell \MakeLowercase{\textit{et al.}}: Bare Demo of IEEEtran.cls for IEEE Journals}

\maketitle

% \begin{abstract}
% The abstract goes here.
% \end{abstract}

% % Note that keywords are not normally used for peerreview papers.
% \begin{IEEEkeywords}
% IEEE, IEEEtran, journal, \LaTeX, paper, template.
% \end{IEEEkeywords}

\input{contents/abstract}

% \fi
\IEEEpeerreviewmaketitle

\input{contents/introduction}
\input{contents/related_work}
\input{contents/problem_formulation}

\input{contents/algorithm}
\input{contents/experiment}
\input{contents/conclusion}
\input{contents/appendix}

\bibliographystyle{IEEEtran}
\bibliography{cite}

\end{document}

%% file: contents/abstract.tex
\begin{abstract}
 Modern meta-reinforcement learning (Meta-RL) methods are mainly developed based on model-agnostic meta-learning, which performs policy gradient steps across tasks to maximize policy performance. However, the gradient conflict problem is still poorly understood in Meta-RL, which may lead to performance degradation when encountering distinct tasks. To tackle this challenge, this paper proposes a novel personalized Meta-RL (pMeta-RL) algorithm, which aggregates task-specific personalized policies to update a meta-policy used for all tasks, while maintains personalized policies to maximize the average return of each task under the constraint of the meta-policy. We also provide the theoretical analysis under the tabular setting, which demonstrates that the convergence of our pMeta-RL algorithm. Moreover, we extend the proposed pMeta-RL algorithm to a deep network version based on soft actor-critic, making it suitable for continuous control tasks. Experiment results show that proposed algorithms outperform other previous Meta-RL algorithms on Gym and MuJoCo suites.
\end{abstract}


%% file: contents/introduction.tex
\section{Introduction}
\label{submission}
Reinforcement learning (RL)~\cite{sutton1988learning, watkins1992q} has long been an interesting research topic under the development of artificial intelligence, which has demonstrated extraordinary capabilities in many fields, such as playing games \cite{mnih2015human, mnih2016asynchronous} and robot control \cite{levine2016end, lillicrap2015continuous}. 
Despite its successful applications, deep RL still suffers from data inefficiency when training agents in specific environments. 
And the learned policy may be overfitted and cannot be well generalized to other unseen tasks.

Meta-learning, also known as learning to learn, has recently gained increasing attention for its success in improving sample efficiency in regression, classification and RL tasks. 
Unlike meta-supervised learning, meta reinforcement learning (meta-RL) needs to automatically extract prior knowledge from previous tasks and achieve fast adaptation \cite{thrun2012learning}. Mainstream meta RL methods are based on model-agnostic meta-learning algorithms, such as MAML \cite{finn2017model}, E-MAML \cite{stadie2018some}, which update model parameters from gradients across tasks by discriminating the learning process and adaptation. However, these methods do not consider the problem of task diversity. 

Some other Meta-RL algorithms attempt to learn an inference network from historical transition samples of different tasks during the meta-training process and build a mapping from observation data to task-relevant information \cite{rakelly2019efficient,zintgraf2019varibad,zhang2021metacure}. Then the agent can distinguish the current task according to the inference network and perform action selection to alleviate the gradient conflict problem. However, these methods jointly learn diverse robot manipulation tasks with a shared policy network, which may hurt the final performance compared to independent training in each task~\cite{yu2020meta}. A major reason is that it is unclear how tasks will interact with each other when jointly trained, and optimizing some tasks may negatively affect others \cite{teh2017distral}.

To tackle this problem, in this paper, we propose a personalized meta reinforcement learning (pMeta-RL) framework without sharing trajectories.
More specifically, we let each task train its own policy and obtain a meta-policy for multiple tasks by sharing policy model parameters and aggregation. In order for each task to have its own personalized policy while contributing to the meta-policy, we introduce a personalization constraint in the optimization objective function to associate the meta-policy and the personalized policy. Furthermore, we extend the framework to deep RL scenario and employ alternating optimization methods to solve the problems. Under the constraint of the meta-policy, each task updates the personalized policy according to its corresponding transition samples, and maintains an auxiliary policy that can be used for meta-policy learning.

\subsection{Main Contributions}
First, we propose a personalization framework to improve the performance of meta-RL on distinct tasks, which learns a \emph{meta-policy} and \emph{personalized policies} for all tasks and specific tasks, respectively. In particular, by learning multiple task-specific personalized policies, the performance on each task can be improved. Meanwhile, to aggregate these differentiated personalized policies to obtain a meta-policy applicable to all tasks, our framework addresses the gradient conflict problem by adopting personalization constraint in the objective function. Under the tabular setting, we constrain the meta Q-table and personalized Q-tables by formulating a differentiable function to encourage each task to pursue its personalized Q-tables around the meta Q-table. Under the deep RL setting, we obtain the personalization performance by constraining the personalized and meta networks (e.g., actor, critic, and inference networks), so that the networks can refer to each other during training.

Second, we propose an alternating minimization algorithm, named pMeta-RL, to solve the personalized meta-RL problem. The personalized policies are updated based on the value iteration, while the \emph{auxiliary policies} used to aggregate the meta-policy are updated based on the gradient descent. More importantly, theoretical analysis shows that the proposed algorithm can converge well, and the convergence speed is basically linear with the iteration number. And we give an upper bound on the difference between the personalized policies and meta-policy, which is influenced by the regularization parameter and task diversity. Moreover, we extend the proposed pMeta-RL algorithm to a deep network version (named deep pMeta-RL) based on soft actor-critic (SAC), making it suitable for continuous control tasks.

Finally, we evaluate the performance of pMeta-RL and deep pMeta-RL on the Gridworld environment and the MuJoCo suites, respectively. 
The experimental results show that the proposed pMeta-RL can achieve better performance than the model averaging method, and verify the conclusion of the convergence analysis. Compared to other previous Meta-RL algorithms, our deep pMeta-RL algorithm can achieve the highest average return on the MuJoCo  locomotion tasks.

%% file: contents/related_work.tex
\section{Related Works}
\subsection{Meta Reinforcement Learning:}
% Meta-learning, also known as learning to learn, has recently gained increased attention for successfully improving sample efficiency in regression, classification and RL tasks.
Existing meta-RL methods learn the dynamics models and policies that can quickly adapt to unseen tasks, which is built on the meta-learning framework in the context of RL \cite{thrun2012learning, duan2016rl, finn2017model}.
$\text{RL}^2$ \cite{duan2016rl} formulate the meta-learning problem as a second
RL procedure, it represent a ``fast'' RL algorithm as a recurrent neural network and learn it from data but update the parameters at a ``slow'' pace. MAML \cite{finn2017model} meta-learns 
model parameters by differentiating the learning process for fast adaptation on unseen tasks. 
E-MAML and E-$\text{RL}^2$ \cite{stadie2018some} explicitly optimize the per-task sampling distributions during adaptation with respect to the expected future returns, which is closely related to the MAML algorithm. ProMP \cite{rothfuss2018promp} theoretically analyses the MAML formulation and addresses the biased meta-gradient issue. To achieve sample efficiency, PEARL \cite{rakelly2019efficient} performs structured exploration by reasoning about uncertainty over tasks and enables fast adaptation by accumulating experience online. VariBAD \cite{zintgraf2019varibad} learns a policy that conditions on this posterior belief, which can  trade off exploration and exploitation under task uncertainty. Other meta-RL works, such as MAME \cite{gurumurthy2020mame} and MetaCURE \cite{zhang2021metacure}, learn a separate exploration policy for general dense and sparse-reward tasks.

\subsection{Multi-task Reinforcement Learning:} 
Multi-task RL methods have demonstrated that learning with different tasks can benefit each other in robotics and other fields \cite{pinto2016curious, pinto2017learning, riedmiller2018learning}. However, the gradient conflict problem still exists in multi-task RL, various approaches based on compositional models \cite{rosenbaum2017routing, d2019sharing, vuong2019sharing, yang2020multi}, gradients similarity \cite{yu2020gradient, chen2018gradnorm}, or policy distillation \cite{parisotto2015actor, teh2017distral, xu2020knowledge} have been proposed to solve this challenge.
In particular, multi-head multi-task SAC \cite{haarnoja2018soft} \cite{yu2020meta} and experience sharing networks \cite{d2019sharing, vuong2019sharing} training with diverse robot manipulation tasks jointly with a sharing network backbone and multiple task-specific heads for actions. 
Hard routing \cite{rosenbaum2017routing} and Soft modularization \cite{yang2020multi} learn a routing network to reconfigure different modules for different tasks automatically without explicitly specifying the policy structure, which can avoid the gradients conflicts effectively.
Other works such as gradient surgery \cite{yu2020gradient} and normalization \cite{chen2018gradnorm}, leverage the similarity between gradients from different tasks to enhance the learning process and mitigate interference. 
Existing algorithms have shown superior performance in multiple tasks,
however, they need to share trajectories between tasks and are difficult to generalize to unseen tasks.

\subsection{Federated Reinforcement Learning:} 
Another related work is the federated reinforcement learning (FRL), these methods encourage multiple agents to federatively build a better decision-making policy in a privacy-preserving manner, which have been applied in robot navigation and Internet of things \cite{liu2019lifelong,wang2020federated,fan2021fault}. \cite{wang2020federated} analyzes the communication overhead in FRL and gives the convergence bound. 
\cite{fan2021fault} studies adversarial attacks in FRL systems in detail and gives theoretical guarantees in the event of participant failure.
However, exist FRL algorithms usually only consider the same tasks and lack the ability to handle multiple tasks.

%% file: contents/problem_formulation.tex
\begin{figure*}[!htbp]
	\centering
	\subfloat{
	\includegraphics[width=7in]{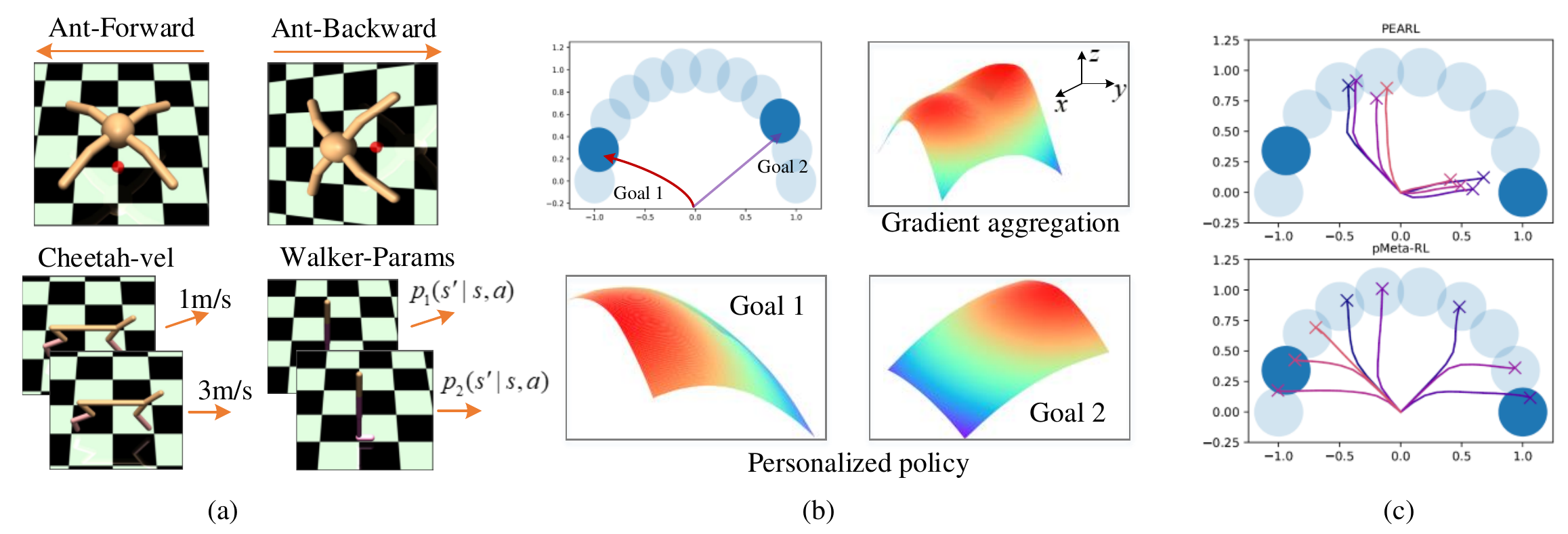}
	}
	\caption{(a) Several distinct tasks on MuJoCo. The tasks vary in either the transition function (e.g., robots with different dynamics) or the reward function (e.g., different goal location).
	(b) Visualization of 2D Sparse-Point-Robot. 
	The blue circles represent the different tasks.
	We highlight two distinct goals in dark blue and visualize the policies leading the agent to reach them.
	$x$ and $y$ axes of the heatmap represent the two dimensions of action, while $z$ axis represents the state-action value on the initial state. Red represents higher values.
	For goals 1 and 2, the personalized policies prefer left and right actions, respectively. In contrast, traditional gradient aggregation-based meta-policy have similar propensities for left and right actions.
	(c) Performance comparison of our algorithm and PEARL at 50000 training steps. 
	The agent using the personalized policy can navigate to more right goals.}
	\label{fig_system}
\end{figure*}

\section{Problem Formulation}
\subsection{Preliminaries}

Reinforcement learning (RL) algorithms are designed to solve sequential decision problems~\cite{watkins1992q,sutton1988learning, mnih2015human}. Generally speaking, the agent makes action decisions at each discrete time step based on its observation of the environment $\mathcal{E}$. After performing an action, the environment generates a reward and transfers the agent to a new state. This can be formulated as a standard Markov decision process (MDP), consisting of a tuple $\mathcal{G}=\left \langle\mathcal{S}, \mathcal{A}, \mathcal{P}, \mathcal{R}, \gamma \right \rangle$, where $\mathcal{S}$ is a finite set of states, $\mathcal{A}$ is a finite set of actions, $r \in \mathcal{R}(s, a): \mathcal{S} \times \mathcal{A} \rightarrow \mathbb{R}$ is the bounded reward function with $\mathcal{R}(s,a) \leq \mathcal{R}_{\max}$, $\mathcal{P}(s^{\prime}\left|s\right., a): \mathcal{S} \times \mathcal{A} \times \mathcal{S} \rightarrow \left[0,1\right]$ denotes the state transition function, and $\gamma \in \left[0,1\right)$ is the discount factor. The goal of RL is to find the optimal policy $\pi^{*}$ by maximizing the expected long-term return.

\subsection{Conventional Meta-RL methods}
Most gradient-based meta-RL algorithms are model-agnostic meta-learners built upon MAML, performing vanilla policy gradient steps towards maximizing the performance of the policy for each task.
These methods generally consider tasks drawn from a distribution $\mathcal{T}_i \sim p\left(\mathcal{T}\right)$, where each task $\mathcal{T}_i$ is a different MDP.
During the meta-training time, the goal of Meta-RL is to learn a policy $\pi(a|s)$, which can adapt to the multiple tasks. In particular,
this policy $\pi(a|s)$ can be optimized by maximizing the average expected return across the task distribution,
\begin{equation}\label{Meta-RL}
\begin{split}
    \max_{\pi} \left\{\mathbb{E}_{\mathcal{T}_i \sim p(\mathcal{T})} \mathbb{E}_{\substack{s \sim \rho_0 \\ a \sim \pi}} [Q^{\pi}_{i}(s,a)]
    \right\},
\end{split}
\end{equation}
where $\rho_0$ is the initial state-distribution, and $Q^{\pi}_{i}(s,a)$ is the state-action value function.

\subsection{Towards Personalized Meta-RL}
Existing meta-RL algorithms leverage the transitions from similar tasks to improve the sample efficiency and achieve good performance on future new tasks.
However, most gradient-based methods suppose that the task distribution is ``homogeneous'', which means that all tasks are from the same domain, (i.e. a single dataset, multiple datasets with same input feature space \cite{vuorio2019multimodal} or robot control with different transition functions but similar dynamics \cite{finn2017model}).
This limits the application of these algorithms to \emph{distinct} or ``hetergeneous'' tasks, i.e., different tasks may vary in environments or transition functions with a large interval \cite{chen2021hetmaml}.
Figure \ref{fig_system}(a) shows several distinct tasks, such as navigating to two completely opposite directions on Ant robot or controlling the speed of the Half-Cheetah to reach two endpoints of a speed interval.
In this setting, optimizing some tasks in meta-RL algorithms may negatively affect others. 

Figure \ref{fig_system}(b) illustrates these problem. 
% Figure \ref{fig_system}(b) illustrates the gradient conflict problem. 
For distinct tasks, gradient aggregation makes the values of different state-action pairs similar, i.e., a recommended action may not be given accurately.
Other Meta-RL methods, such as PEARL \cite{rakelly2019efficient}, distinguish different tasks by learning a task-relevant context $\mathbf{z}$ based on the history. 
These algorithms maintain an inference network $q_{\zeta}(\mathbf{z}|\mathbf{c})$, parameterized by $\zeta$, to encode salient information about tasks from context $\mathbf{c}$, which consists of the transitions of different tasks. Nevertheless, the gradient conflict problem still exists due to the shared policy network parameters.
To solve this challenge, we aim to learn a \emph{meta-policy} that adapts to multiple distinct tasks, while optimizing a \emph{personalized policy} for each task.
In this subsection, we first give the definition of personalized Meta-RL under the tabular and deep network settings, while the corresponding algorithm for solving personalized Meta-RL is proposed in the next section.

\begin{definition}[Personalized Meta-RL] \label{def1} 
Assume that there are $N$ tasks and the $i$-th ($i = 1,...,N$) task $\mathcal{T}_i$ drawn from $p(\mathcal{T})$. 
Define $Q^{\pi}(s,a), s\in {\mathcal{S}}, a \in {\mathcal{A}}$ as the \emph{meta Q-table}, where 
${\mathcal{S}} =  \cup_{i\in N} \mathcal{S}_i$ and ${\mathcal{A}} = \cup_{i\in N} \mathcal{A}_i $ with $\mathcal{S}_i$ and $\rho_{i,0}$ being the set of the $i$-th task's states and initial state distribution, respectively. The personalized Meta-RL aims to train a \emph{meta-policy} $\pi$ by maximizing
\begin{equation}\label{pFedRL1}
    \max_{\pi} \Bigg\{ \mathcal{L}(\pi) := \frac{1}{N}\sum_{i=1}^{N} \mathcal{L}_i(\pi)
    \Bigg\},
\end{equation}
where 
\begin{align}
    \mathcal{L}_i(\pi) &=  \max_{\pi_i}   L_i(\pi_i)
    - \frac{\lambda}{2} \varphi_i(Q^{\pi_i}_i,Q^{\pi}), \nonumber \\
    L_i(\pi_i) &= \mathbb{E}_{\substack{s_i \sim \rho_{i,0}\\a_i \sim \pi_i}} \sum_{t=0}^{\infty}\gamma^t\mathcal{R}_{i}(s_{i,t}, a_{i,t})|s_{i,0}=s_i, s_{i,t}, a_{i,t}  \nonumber 
\end{align}
with $\lambda$ being a weight parameter that controls the level of personalization; $Q^{\pi_i}_i(s_i,a_i)$ being the personalized Q-table of the $i$-th task; $L_i(\pi_i)$ being the expected long-term return under the policy $\pi_i$; and  $\varphi_i(\cdot)$ being a differentiable constraint function, such as $\ell_2$-norm.
\end{definition}

Definition~\ref{def1} describes a  personalization approach based on Q-learning, where the personalized constraint enables each task optimizes its own policy around $Q^{\pi}(s,a)$. 
Since Q learning is difficult to adapt to large-scale problems, we use a function approximator to represent the Q-values, and extend Definition~\ref{def1} to deep personalized meta-RL based on SAC \cite{haarnoja2018soft} and inference network $q_{\zeta}$.

\begin{definition}[Deep Personalized Meta-RL] \label{def2} 
Assume that there are $N$ training tasks drawn from $p(\mathcal{T})$. Define $\pi_{\omega}(a|s,\mathbf{z})$ as the \emph{meta-policy}, where $\mathbf{z}$ is the task-relevant vector.
Deep personalized meta-RL aims to train a meta actor model $\pi_{\omega}$, a meta Q function $Q_{\vartheta}$ and a meta inference network $q_{\varpi}$ by alternately minimizing
\begin{align}
    \min_{\omega \in \mathbb{R}^d} F(\omega)&:=\frac{1}{N}\sum_{i=1}^N  F_{i,\pi}(\omega), \nonumber \\
    \min_{\vartheta \in \mathbb{R}^d}    F(\vartheta)&:=\frac{1}{N}\sum_{i=1}^N  F_{i,Q}(\vartheta), \nonumber \\
    \min_{\varpi \in \mathbb{R}^d} F(\varpi) &:= \frac{1}{N}\sum_{i=1}^N F_{i,q}(\varpi), \nonumber
\end{align}
where
\begin{align}
    F_{i,\pi}(\omega) =&~ \min_{\phi_i\in\mathbb{R}^d} J_{\pi_i}(\phi_i)+\frac{\lambda}{2} \varphi_i(\phi_i, \omega), \nonumber \\
    F_{i,Q}(\vartheta) =&~ \min_{\theta_{i}\in\mathbb{R}^d} J_{Q_i}(\theta_{i})+\frac{\lambda}{2} \varphi_{i}(\theta_{i}, \vartheta), \nonumber\\
    F_{i,q}(\varpi) =&~ \min_{\zeta_{i}\in\mathbb{R}^d} J_{Q_i}(\theta_{i}) + \frac{\lambda}{2} \varphi_{i}(\zeta_{i}, \varpi) \nonumber  \\
    &+ \mathbb{E}_{q_{\zeta}(\mathbf{z}|\mathbf{c}^{\mathcal{T})}}\left[D_{\text{KL}}(q_{\zeta}(\mathbf{z}|\mathbf{c}^{\mathcal{T})} \big\| p(\mathbf{z}))\right] \nonumber 
\end{align}
with $\phi_i$, $\theta_{i}$ and $\zeta_i$ being the personalized actor model, soft Q function model and inference model of the $i$-th task respectively, $J_{\pi_i}(\phi_i)$ and $J_{Q_i}(\theta_{i})$ being the actor and critic objective function, which is given by
\begin{equation}
    J_{\pi_i}(\phi_i) = \mathbb{E}_{\substack{s_t \sim \mathcal{D}_i \\ a_t\sim \pi_{\phi_i}}}\left[\alpha \log \pi_{\phi_i}(a_t|s_t, \mathbf{z}) - Q_{\theta_i}(s_t, a_t, \mathbf{z})\right],\nonumber 
\end{equation}
and 
\begin{align}
    J_{Q_i}(\theta_i) &= \mathbb{E}_{\substack{(s_t, a_t) \sim \mathcal{D}_i \\ \mathbf{z}\sim q_{\zeta_i}(\mathbf{z}|\mathbf{c})}} \Big[\frac{1}{2}\big(Q_{\theta_i}(s_t, a_t, \mathbf{z}) \nonumber \\ 
    &~~~~-(r(s_t, a_t)+\gamma \mathbb{E}_{s_{t+1} \sim p} \left[V_{\bar{\theta_i}}(s_{t+1}, \bar{\mathbf{z}})\right])\big)^2\Big], \nonumber
\end{align}
respectively, 
where $\bar{\mathbf{z}}$ indicates that gradients are not being computed through it. 
\end{definition}

Definition~\ref{def2} considers the personalization approach to three models $(\phi_i, \theta_i, \zeta_i)$,
and establishes the relationship between them to optimize the personalized policy based on the meta-policy. In the next section, we will propose the corresponding algorithms to solve these problems.

%% file: contents/algorithm.tex
\section{Proposed Algorithms}\label{A3}
In this section, we first propose the pMeta-RL algorithm to solve personalized Meta-RL in Definition~\ref{def1} and present the corresponding theoretical analysis. 
Then we extend this algorithm with deep neural network to solve deep personalized Meta-RL in Definition~\ref{def2}.
\subsection{Personalized Multi-task Value Iteration}\label{A31}
To solve personalized Meta-RL, our pMeta-RL algorithm updates $Q_i^{\pi_i}$ and $Q^{\pi}$ by alternately minimizing the two subproblems of pMeta-RL. More specifically, we first update $Q_i^{\pi_i}$ by solving
\begin{align}\label{eq_sub_1}
    \hat{Q}_i^{\pi_i} = \arg \min_{Q_i^{\pi_i}} -L_i(\pi_i)
    + \frac{\lambda}{2} \varphi_i(Q^{\pi_i}_i,Q^{\pi}),
\end{align}
then we introduce an auxiliary meta Q-values $Q_i^{\pi}$ for each task, which can be updated by solving
\begin{align}\label{eq_sub_2}
    \hat{Q}_i^{\pi} = \arg \min_{Q_i^{\pi}} 
    \frac{\lambda}{2} \varphi_i(Q^{\pi_i}_i,Q_i^{\pi}).
\end{align}
After that, all auxiliary meta Q-values $Q_i^{\pi}, i = 1, \cdots, N$ can be aggregated to update the $Q^{\pi}$.

We first focus on solving \eqref{eq_sub_1}. 
Note that in the model-free RL, the standard Q-learning can be used to solve $\arg \min_{\pi_i}-L_i(\pi_i)$. For each state-action pair $(s,a)$, $Q_i^{\pi_i}(s,a)$ is estimated by iteratively applying the Bellman optimal operator $\mathcal{B}^*\left[Q_i^{\pi_i}(s,a)\right] = \mathcal{R}_i(s,a)+\gamma \mathbb{E}_{s^{\prime}\sim \mathcal{P}(s^{\prime}|s,a)}\left[\max_{a^{\prime}}Q_i^{\pi_i}(s^{\prime},a^{\prime})\right]$, i.e.,
$$Q^{\pi_i}_{i,k+1}(s,a) \leftarrow \arg \min_{Q_i^{\pi_i}} \frac{1}{2} \left\|Q_i^{\pi_i}(s, a) - \mathcal{B}^*\left[Q^{\pi_i}_{i,k}(s,a)\right] \right\|_2^2 ,$$
where $Q^{\pi_i}_{i,k}(s,a)$ is the Q-table at the $k$-th iteration. Then the exact or an approximate maximization scheme is used to recover the greedy policy. This inspires us to update $Q^{\pi_i}_{i}(s_i,a_i)$ by performing the following iterations
\begin{equation}\label{pFedRL-A1}
    Q^{\pi_i}_{i,k+1}(s_i,a_i) \leftarrow \arg \min_{Q^{\pi_i}}
    \Xi_{i,k}(Q^{\pi_i}),
\end{equation}
where
\begin{align}
\Xi_{i,k}(Q^{\pi_i}) =&~ \frac{1}{2} \left\| Q^{\pi_i}(s_i,a_i) - \mathcal{B}^*[Q^{\pi_i}_{i,k}(s_i,a_i)] \right\|_2^2 \nonumber \\
    &~~~~~~~~~~~~~~~~+ \frac{\lambda}{2}\varphi_i (Q^{\pi_i}_i(s_i,a_i), Q^{\pi}_{i}(s_i,a_i)). \notag
\end{align}
We solve \eqref{pFedRL-A1} via one step gradient descent based on the gradient $\nabla \Xi_{i,k}\left(Q_i^{\pi_i}\right)$ as the following
\begin{align}\label{per_update_eq}
    & Q_{i,k+1}^{\pi_i}(s_i,a_i) = Q_{i,k}^{\pi_i}(s_i,a_i) + \eta^k \Big[ \mathcal{R}_{i}(s_i,a_i) +    \\
    & \gamma \mathbb{E}_{s_i^{\prime}\sim \mathcal{P}_i}
    \{\max_{a_i^{\prime}} Q_{i,k}^{\pi_i}(s_i^{\prime},a_i^{\prime})\}
     - Q_{i,k}^{\pi_i}(s_i,a_i)\Big] + \frac{\eta^k \lambda}{2} \nabla \varphi_i, \nonumber
\end{align}
where $\nabla \varphi_i = 2\left(Q_i^{\pi_i}(s_i,a_i)-Q_i^{\pi}(s_i,a_i)\right)$ and $\eta^k$ is a learning rate. However, it is not straightforward to update $Q_i^{\pi_i}$ based on \eqref{per_update_eq} since the system transition function $\mathcal{P}_i$ is usually unknown. In most model-free RL algorithms, we need to sample transition data to update Q-value. We hence leverage several samples $(s_i,a_i,r_i,s_i^{\prime})$ to obtain an approximated Q-table $\widetilde Q_{i}^{\pi_i}$ as the following, 
\begin{align}\label{per_update0}
    &\widetilde Q_{i,k+1}^{\pi_i}(s_i,a_i) = \widetilde Q_{i,k}^{\pi_i}(s_i,a_i) + \eta^k \Big\{ \mathcal{R}_{i}(s_i,a_i)+   \nonumber \\
    &~~~~~~~\gamma  \max_{a_i^{\prime}} \widetilde Q_{i,k}^{\pi_i}(s_i^{\prime},a_i^{\prime})
    - \widetilde Q_{i,k}^{\pi_i}(s_i,a_i)\Big\} + \frac{\eta^k \lambda}{2} \nabla \varphi_i.
\end{align}
The iterations in \eqref{per_update_eq} and \eqref{per_update0} go until the maximum iteration number $K$ is reached. Then we have the following Theorem~\ref{delta_converge}, which is proved in Appendix~\ref{proof1}. 
Theorem~\ref{delta_converge} indicates that the error generated by using the approximated Q-table $\widetilde Q_{i}^{\pi_i}$ is minimal after sufficient iterations. We hence set $ Q_{i}^{\pi_i}(s_i,a_i) \leftarrow \widetilde Q_{i,K}^{\pi_i}(s_i,a_i)$.
\begin{theorem}\label{delta_converge}
     For a large enough $K$, there exists a small $\delta$ satisfies
     \begin{equation}
         \left| Q_{i,K}^{\pi_i}(s_i,a_i) - \widetilde Q_{i,K}^{\pi_i}(s_i,a_i)\right| \leq \delta, \forall s_i, a_i, i. \nonumber
     \end{equation}
\end{theorem}

Once $Q_i^{\pi_i}(s_i, a_i)$ is available, we update the auxiliary Q values $Q_i^{\pi}$ in \eqref{eq_sub_2} as the following
\begin{align}\label{local_update0}
    Q_{i}^{\pi}(s_i, a_i) =&~ Q_{i}^{\pi}(s_i, a_i) - \eta \nabla \mathcal{L}_i( Q_{i}^{\pi}(s_i, a_i) ) \\
    =&~ Q_{i}^{\pi}(s_i, a_i) + \eta \lambda \big[
    Q_{i}^{\pi_i}(s_i, a_i)- Q_{i}^{\pi}(s_i, a_i)
    \big], \nonumber
\end{align}
where $\eta$ is a learning rate.

We solve these two subproblems alternately until the maximum number $R$ of iterations is reached and then update the meta Q-values.
Note that since $s\in {\mathcal{S}} =  \cup_{i\in N} \mathcal{S}_i, a \in {\mathcal{A}} = \cup_{i\in N} \mathcal{A}_i $, the states and actions in the meta Q-table may not necessarily be in a specific personalized Q-table. Hence all the auxiliary meta Q-tables $Q_{i, R}^{\pi}(s_i, a_i), i = 1,...,N$ are collected to perform the following aggregation
\begin{equation}\label{global_update0}
\begin{split}
     \hat{Q}^{\pi}(s, a) &= (1-\beta) Q^{\pi}(s, a) \\
    &~~~~~~~~ +  \frac{\beta}{N_s} \sum_{i=1}^{N}  Q_{i}^{\pi} \cdot \mathbb{I}\{ Q_{i}^{\pi} (s, a) \neq \emptyset \},
\end{split}
\end{equation}
where $\beta$ is a weight parameter, $\mathbb{I}\{\cdot\}$ is an indicator operator, and $N_s$ denotes the number of tasks that satisfy $Q_{i}^{\pi}(s, a) \neq \emptyset$, i.e., $|\mathbb{I}\{ Q_{i}^{\pi} (s, a) \neq \emptyset \}|$. We repeat the above process until the meta-policy converges or reaches the maximum number of iterations $C$.

\subsection{Theoretical Analysis}\label{A32}
In this subsection, we present the convergence analysis of our pMeta-RL algorithm.
First, we present the following assumption, which quantifies the diversity of the reward and transition function among tasks.
\begin{assumption}\label{assumption_diversity}
For any state-action pair $(s,a)$, the reward function of each task can be bounded by $\mathcal{R}_i(s,a)-\frac{1}{N}\sum_{j=1}^{N}(\mathcal{R}_j(s,a)) \leq \sigma_{1,i}$, while the transition function is bounded by $\mathcal{P}_i(s^{\prime}|s,a)-\frac{1}{N}\sum_{j=1}^{N}(\mathcal{P}_j(s^{\prime}|s,a)) \leq \sigma_{2,i}$.
\end{assumption}

Then we have the following Theorem \ref{theorem_converge} based on Assumption~\ref{assumption_diversity}, which is proved in Appendix~\ref{proof2}.
\begin{theorem}\label{theorem_converge}
    Let Assumption \ref{assumption_diversity} holds. When $\eta \leq \frac{\tilde{\eta}}{\beta R}$ and
    \begin{equation}
        \tilde{\eta}L\left(\frac{3}{2}+\frac{12}{\lambda^2-8}+ \frac{24\lambda^2}{\lambda^2-8}\right) \leq \frac{1}{4}, \nonumber
    \end{equation}
we have
    \begin{align}
            (a)~&~ \mathbb{E}\left[\left\|\nabla \mathcal{L}(Q^{\pi, *})\right\|^2\right] \leq \mathcal{O}\Bigg(\frac{\Delta}{\hat{\eta}_{2} C}+  \nonumber\\
            &\frac{\Delta^{\frac{2}{3}}\left(96L^2(\lambda^2\delta^2+\sigma_2^2)\right)^{\frac{1}{3}}}{\left(\beta C\right)^{\frac{2}{3}}}+
            \frac{\left(\Delta 3L\sigma_2^2\right)^{\frac{1}{2}}}{\sqrt{C}}+3\lambda^2\delta^2\Bigg), \nonumber \\
            (b)~&~ \sum_{i=1}^{N} \mathbb{E}\left[\left\|\hat{Q}_{i,c}^{\pi_i}-Q_{c}^{\pi}\right\|^{2}\right] \leq \nonumber \\
            &~~~~\mathcal{O}\left(\mathbb{E}\left[\left\|\nabla \mathcal{L}(Q^{\pi,*})\right\|^2\right]\right)
            + \mathcal{O}\left(\delta^2 + \frac{2\sigma_2^2}{\lambda^2}\right), \nonumber
    \end{align}
where $L$ is the smoothness of $\mathcal{L}(\pi)$, $\Delta = \mathcal{L}(Q^{\pi}_0)-\mathcal{L}^{*}$ and $\sigma_2^2=\frac{2\lambda^2}{\lambda^2-8}\sigma^2+\frac{2\lambda^2\gamma}{(\lambda^2-8)(1-\gamma)}\mathcal{R}_{\text{max}}$ with $\sigma^2=\frac{1}{N}\sum_{i=1}^{N}\left(\sigma_{1,i}+\frac{\sigma_{2,i}\gamma}{(1-\gamma)}\mathcal{R}_{\text{max}}\right)^2$.
\end{theorem}

\begin{remark}
Theorem \ref{theorem_converge}(a) shows the convergence of the meta-policy. The first term is caused by the initial error $\Delta$, which decreases linearly with the increase of training iterations. Moreover, the second and third terms are respectively due to the tasks drift and initial error $\Delta$, which decreases as the increase of training iterations. Finally, the last term shows that the gradient of the optimal policy model can be close to zero when $\delta$ is small enough. 
Theorem \ref{theorem_converge}(b) describes an upper bound on the distance between the personalized policy and the meta-policy. The first term denotes the convergence rate consistent with the meta-policy. The $\sigma^2$ in the second term indicates that the diversity of tasks can increase the upper bound, while a more prominent regularization factor $\lambda$ can strengthen the connection to the meta-policy.
\end{remark}

\subsection{Deep Personalized Meta-Reinforcement Learning}\label{A33}
In this subsection, we extend the meta-policy iteration algorithm to a deep learning version with neural network function approximators. Taking the actor model $\phi_i$ as an example, the deep pMeta-RL updates the personalized policy by solving the following two sub-problems:
\begin{align}
    \hat{\phi}_i &= \arg \min_{\phi_i} J_{\pi_i}(\phi_i)
    + \frac{\lambda}{2} \varphi_i(\phi_i,\omega),  \label{eq_sub_1d} \\
    \hat{\omega}_i &= \arg \min_{\omega_i}
    \frac{\lambda}{2} \varphi_i(\phi_i,\omega_i), \label{eq_sub_2d}
\end{align}
where $\omega_i$ is the auxiliary model, which is used to update the meta-model $\omega$.

To solve \eqref{eq_sub_1d}, similar with Theorem~\ref{delta_converge}, we sample a mini-batch data $D_i$ to obtain an approximated $\nabla J_{\pi_i}(\phi_i)$ by calculating $\mathbb{E}\left[\nabla J_{\pi_i}(\phi_i; D_i)\right]$. Then, the personalized policy is updated based on the stochastic gradient descent by
\begin{equation}\label{per_update}
    \hat{\phi}_{i} = \phi_{i} - \alpha (\nabla J_{\pi_i}(\phi_{i}; D_i)+\frac{\lambda}{2}\nabla \varphi_i(\phi_{i}, \omega_{i})),
\end{equation}
where we use the $\ell_2$-norm constraint such that $\nabla \varphi_i(\phi_{i}, \omega_{i}) = 2(\phi_{i}-\omega_{i})$ and $\alpha$ is the personalized policy learning rate.

For the subproblem \eqref{eq_sub_2d}, $\omega_{i}$ is updated by
\begin{equation}\label{local_update}
\begin{split}
    \hat{\omega}_{i} = \omega_{i} - \eta \nabla F_{i,\pi}(\omega_{i}) = \omega_{i} - \eta \lambda \left(\omega_{i} - \phi_i\right),
\end{split}
\end{equation}
where $\eta$ denotes the auxiliary model learning rate. Moreover, the critic and inference models are also updated similarly, which we omit here for brevity.

After $R$ iterations, the meta-policy can be updated as the following
\begin{equation}\label{global_update}
    \omega^{c+1}=(1-\beta) \omega^{c}+\beta \sum_{i=0}^N \frac{\omega_{i, R}^{c}}{N},
\end{equation}
where $\beta$ is a weight parameter and $c$ represents the current number of update rounds of the meta-policy.
Finally, we obtain a well-trained meta-policy $(\omega^C, \vartheta^C, \varpi^C)$. 
We summarize pMeta-RL and deep pMeta-RL algorithms in Algorithm \ref{alg2} for clarification.

\begin{algorithm}[tb]
\caption{pMeta-RL / Deep pMeta-RL Algorithms} %算法的名字
\hspace*{0.02in} % \hspace*{0.02in}用来控制位置，同时利用 \\ 进行换行
\textbf{Input}: $C, U, \lambda, \eta, \beta, \omega^0, \vartheta^0, \varpi^0$ and a set of training tasks $\left\{\mathcal{T}_i\right\}_{i=1\cdots N}$ from $p\left(\mathcal{T}\right)$.
\begin{algorithmic}[1] %[1] enables line numbers
\FOR{$c = 1, \cdots, C$}
\FOR[\emph{For deep pMeta-RL only}]{each $\mathcal{T}_i$} 
% \emph{For deep pMeta-RL only}
\STATE Initialize context $\mathbf{c}_i = \left\{\right\}$
\STATE Sample $\mathbf{z}\sim q_{\zeta_i}(\mathbf{z}|\mathbf{c}^i)$
\STATE Update $\mathbf{c}_i = \left\{\left(s_j,a_j,s^{\prime},r_j\right)\right\}_{j=1 \cdots T} \sim \mathcal{D}_i$
\ENDFOR
\STATE Set the meta-policy $Q^{\pi} / \left(\omega^c, \vartheta^c, \varpi^c\right) $ as the personalized policy for each task
\FOR{each $\mathcal{T}_i$ parallel}
\FOR{$r = 1, \cdots, R$}
\STATE \textbf{Personalized policy update:} \\ Update $Q_i^{\pi_i}/(\phi_{i,r}^c, \theta_{i,r}^c, \zeta_{i,r}^c)$ by \eqref{per_update0} or \eqref{per_update}
\STATE \textbf{Auxiliary meta-policy update:} \\ Update $Q_i^{\pi}/(\omega_{i,r}^c, \vartheta_{i,r}^c, \varpi_{i,r}^c)$ by
\eqref{local_update0} or \eqref{local_update}
\ENDFOR
\ENDFOR
\STATE Update meta-policy $Q^{\pi} / \left(\omega^c, \vartheta^c, \varpi^c\right)$ by \eqref{global_update0} or \eqref{global_update}
\ENDFOR
\STATE \textbf{Output}: meta-policy $Q^{\pi} / \left(\omega, \vartheta, \varpi\right)$, personalized policies $Q_i^{\pi_i} / \left(\phi_i, \theta_i, \zeta_i\right), i=1,\cdots, N $
\end{algorithmic}
\label{alg2}
\end{algorithm}

\begin{remark}
Algorithm \ref{alg2} integrates two algorithms, pMeta-RL and deep pMeta-RL. For the tabular setting, our pMeta-RL algorithm updates $Q_i^{\pi}$ without using lines 3-7 in Algorithm \ref{alg2}. Note that our personalization method is a plug-and-play module that can directly combine with many RL algorithms (e.g., DQN, SAC). 
For example, each task learns a SAC policy whose actor and critic models are updated according to lines 9-11. The vector $\mathbf{z}$ used to distinguish different tasks can be replaced by a task embedding vector, such as a one-hot vector. Therefore, we can learn a policy (line 15) to solve multiple distinct tasks in a decentralized manner.
\end{remark}

The deep pMeta-RL in Algorithm \ref{alg2} proposes a general meta-training procedure, and the meta-testing phase is consistent with PEARL. By training an inference network, deep pMeta-RL can estimate the state-value function $Q_{\vartheta}(s,a, \mathbf{z})$ under different tasks and generalize to unseen tasks.

%% file: contents/experiment.tex
\section{Experiment Results}
\subsection{Environments Setting}
For the Gym suite \cite{1606.01540}, we consider two classical control environments, ``CartPole'', and ``MountainCar'' \footnote{The IDs of these environments in the OpenAI Gym library are: \emph{CartPole-v0} and \emph{MonutainCar-v0}}.
For each environment, we modify its physical parameters to induce different transitions as different tasks.

\begin{table*}[htbp]
  \caption{The system parameters of different tasks on the Gym suites.}
  \centering
  \scalebox{1.0}{
  \begin{tabular}{cccccccccccc}
    \toprule
    Environments & Parameters & task1  & task2 & task3 & task4 & task5  \\
    \midrule
    \multirow{3}{*}{CartPole}    &{force}         & 20.0     & 1.0     & 10.0     & 10.0     & 10.0      \\
        &{masspole}      & 0.1      & 0.1     & 1.0     & 0.01     & 0.1       \\
        & {lengthpole}    & 0.5      & 0.5     & 0.5     & 0.5     & 0.05      \\

    \midrule
    \multirow{3}{*}{MountainCar} &{force}  & 0.001     & 0.001     & 0.001     & 0.001     & 0.001     \\
    &{gravity}  & 0.0025     & 0.0025     & 0.0025     & 0.0025     & 0.0025     \\
    &{inclination $\zeta$}         & 3.0     & 3.5     & 4.0     & 4.5     & 5.0     \\
    \bottomrule
  \end{tabular}}
  \label{table_cartpole}
\end{table*}

\textbf{CartPole:}
The goal of CartPole is to balance a pole on the top of the cart. The agent can get the position and velocity of the cart, the angle and angular velocity of the pole as observations, which is a four-dimensional continuous space. We can make the cart move left or right as the action space. This task ends when the pole falls over or the cart out of bounds. The system parameters contain the force, the mass of pole (masspole) and the length of pole (lengthpole).

\textbf{MountainCar:}
The goal of MounatianCar is to reach a specific position on a one dimensional track between two mountains. The agent learns to drive back and forth to obtain enough power to allow the car to reach its goal. The states are the position $\dot{p}$ and velocity $\dot{v}$ of the car and the action consist of pushing left, right and no push.
The system transition function can be written as 
\begin{equation}
\begin{split}
    \dot{v}_{t+1} &= \dot{v}_{t}+(a-1)\hat{F}-\hat{G}\cos(\zeta\dot{p}) \\
    \dot{p}_{t+1} &= \dot{p}_{t} +\dot{v}_{t+1}
\end{split}
\end{equation}
where $a=0,1,2$ represent the push left,right and no push, respectively, $\hat{F}$ is the magnitude of the force, $\hat{G}$ is the gravity, and $\zeta$ controls the inclination of the track. We modify $\zeta$ to generate different tasks, which is shown in table~\ref{table_cartpole}.

For the mojuco suite, we consider the same meta-learning domains as \cite{li2019efficient, zhang2021metacure}.

\textbf{Ant-Fwd-Back:} Move forward or backward (2 train tasks, 2 test tasks).

\textbf{Half-Cheetah-Dir:} Move forward or backward (2 train tasks, 2 test tasks).

\textbf{Half-Cheetah-Vel:} Reach a target velocity (10 train tasks, 3 test tasks).

\textbf{Walker-2D-Params:} The agent is randomly initialized with different system dynamics parameters and keep walking (10 train tasks, 3 test tasks).

\begin{table}[!tb]
  \caption{The hperparameters of pMeta-RL on CartPole and MountainCar.}
  \centering
  \scalebox{1.0}{
  \begin{tabular}{cccccccccccc}
    \toprule
    \multirow{2}{*}{Hyperparameters} & ~~~~~~~~~~~~~~~Value &  &                 \\
    \cmidrule(r){2-4}
                          & CartPole       & MountainCar        \\
    \midrule
        {Learning rate}    & 1e-3   & 1e-3              \\
        {Hidden layer size}         & [512 512] & [512 512]         \\
        {Batch size $D$}            & 64 & 64                  \\
        {Replay buffer size}        & 2e5 & 2e5              \\
        {Meta-policy update round $C$}          & 30  & 30                \\
        {Personalized update round $R$}           & 200  & 500              \\
        {Action selector}           & $\epsilon$-greedy & $\epsilon$-greedy \\
        {$\epsilon$-start}          & 0.3 & 0.3                 \\
        {$\epsilon$-finish}         & 0.01 & 0.01                \\
        {Target update interval}    & 350 & 350                 \\
        {Regularization coefficient $\lambda$}                 & 20  & 20\\
        {Discount factor $\gamma$}                 & 0.99  & 0.99\\
    \bottomrule
  \end{tabular}}
  \label{table_cartpole}
\end{table}

\begin{table*}[!tb]
  \caption{The hperparameters of pMeta-RL on the MuJoCo suite.}
  \centering
  \scalebox{1.0}{
  \begin{tabular}{cccccccccccc}
    \toprule
    \multirow{2}{*}{Hyperparameters} & & ~~~~~~~~~~~~~~~~~~~~~Value &                 \\
    \cmidrule(r){2-5}
                          & Ant-Fwd-Back       & Half-Cheetah-Dir  & Half-Cheetah-Vel    & Walker-2D-Params  \\
    \midrule
        {Actor Learning rate}    & 1e-3   & 1e-3   & 1e-3  & 1e-3              \\
        {Critic Learning rate}   & 1e-3   & 1e-3   & 1e-3  & 1e-3              \\
        {Actor Hidden layer size}  & [512 512]  & [512 512]  & [512 512]  & [512 512]   \\
        {Critic Hidden layer size}  & [512 512]  & [512 512] & [512 512]  & [512 512]  \\
        {Batch size $D$}            & 256   & 256  & 256    & 256         \\
        {Replay buffer $\mathcal{D}_i$ size} & 1e5  & 1e5  & 2e4   & 2e4  \\
        {Meta-policy update round $C$}          & 200   & 300    & 100     & 100       \\
        {Personalized update round  $R$}           & 5000  & 4000  & 2000    & 2000        \\
        {Total timesteps}           & 2e6     & 2.4e6    & 2e6    & 4e6        \\
        {Regularization coefficient $\lambda$}     & 15  & 15  & 20     & 20           \\
        {Target smoothing coefficient} & 0.005 & 0.005 & 0.005 & 0.005 \\
        {Discount factor $\gamma$}     & 0.99  & 0.99  & 0.99 & 0.99  \\
    \bottomrule
  \end{tabular}}
  \label{table_mujoco}
\end{table*}

\subsection{Experiments Details}
We implement our personalized method based on DQN with a prioritized experience replay buffer, all the details can be found in Table~\ref{table_cartpole}. Moreover, our code is based on \emph{Tianshou} \cite{weng2021tianshou} framework\footnote{https://github.com/thu-ml/tianshou}, and all experiments were conducted on a NVIDIA Quadro RTX 6000 environment.

Then we present all the hyperparameter settings in the Gym and MuJoCo tasks. Table~\ref{table_cartpole} shows the hyperparameters used on CartPole and MountainCar. All hyperparameters are basically set to be the same as in \cite{anschel2017averaged}. Table~\ref{table_mujoco} shows the hyperparameters used on MuJoCo suite, where all hyperparameters are basically set to be the same as in \cite{rakelly2019efficient, zhang2021metacure}. For all the compared algorithms, we set the number of training steps to be the same as that used in our pMeta-RL. The rest of the hyperparameters in their algorithm are consistent with the settings recommended in their papers \cite{rakelly2019efficient, zhang2021metacure}.

\subsection{Performance of pMeta-RL: A Warm-Up}

\begin{figure}[!tbp]
	\centering
% 	\hspace{-0.3cm}
	\subfloat{
	\includegraphics[width=1.55in]{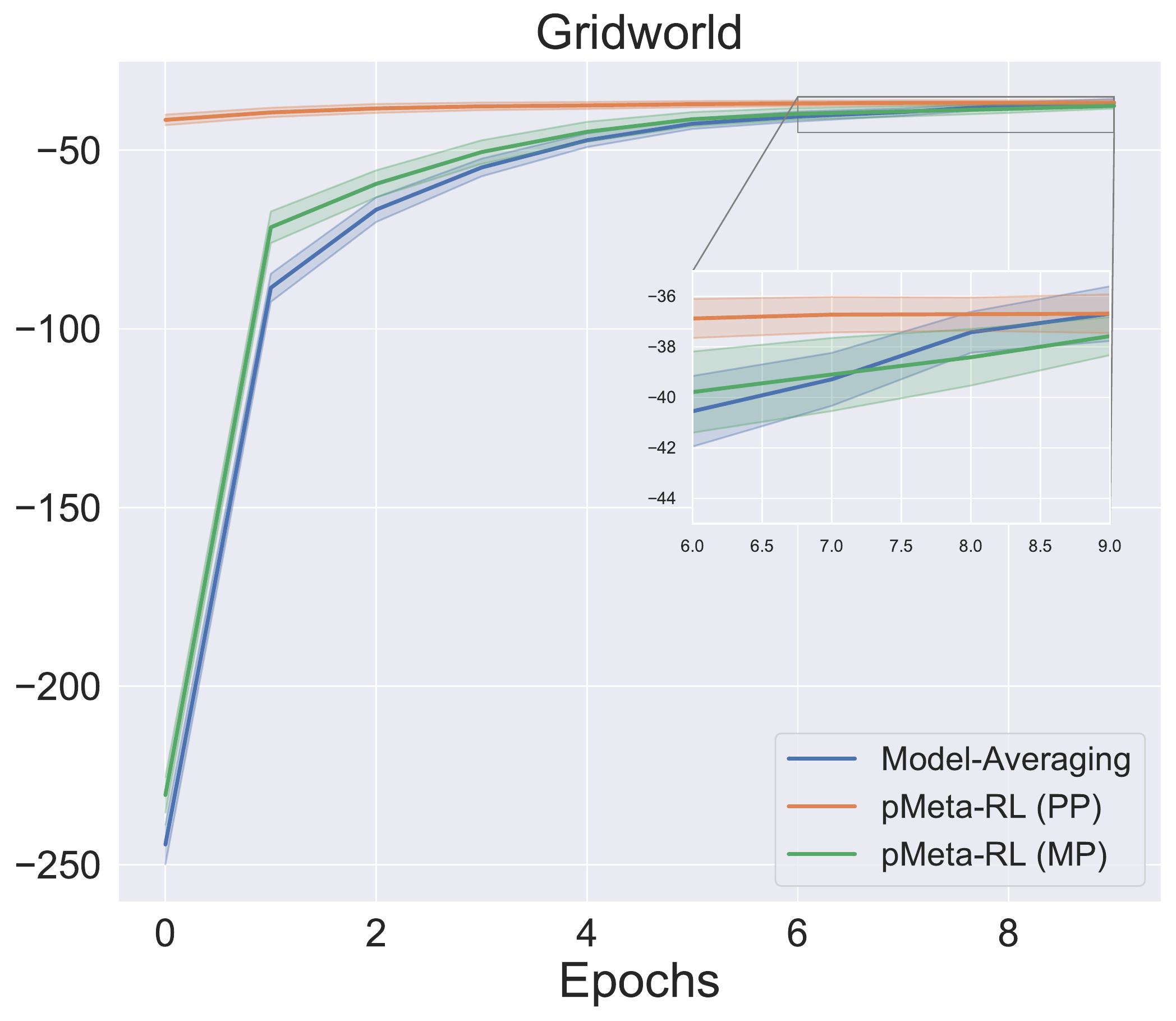}
	}
	\subfloat{
	\includegraphics[width=1.55in]{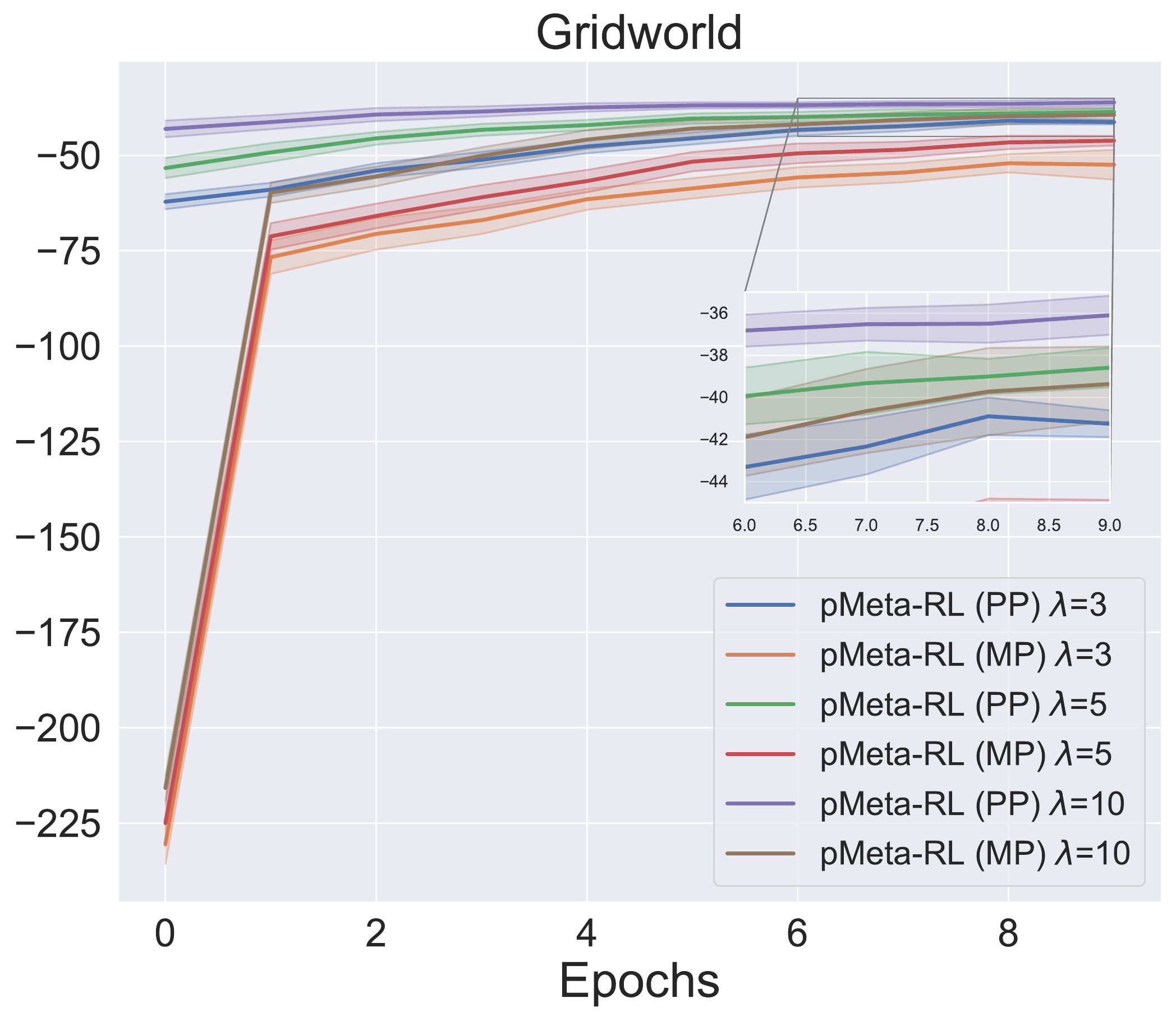}
	}

    % \hspace{-0.3cm}
	\subfloat{
	\includegraphics[width=1.55in]{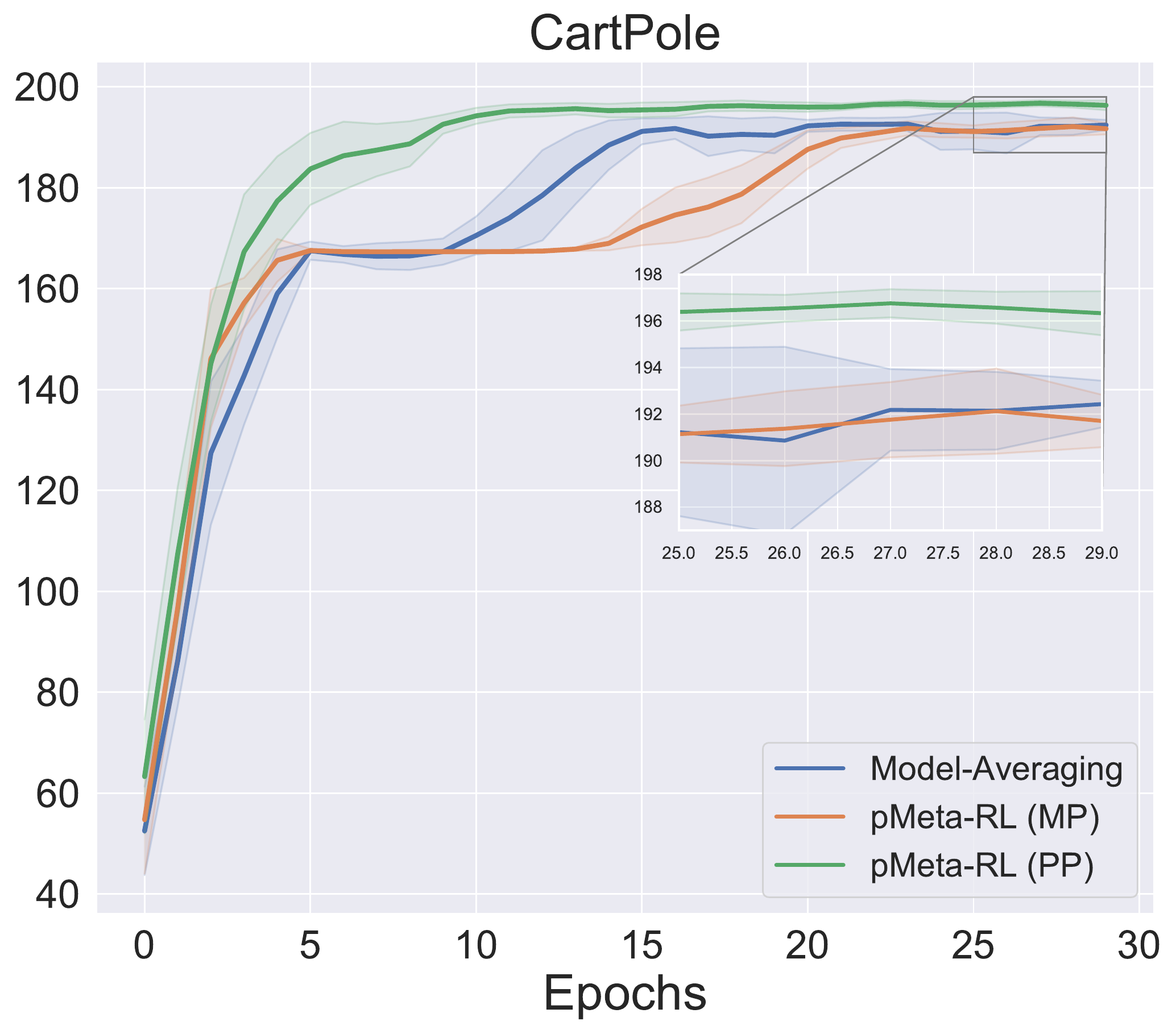}
	}
	\subfloat{
	\includegraphics[width=1.55in]{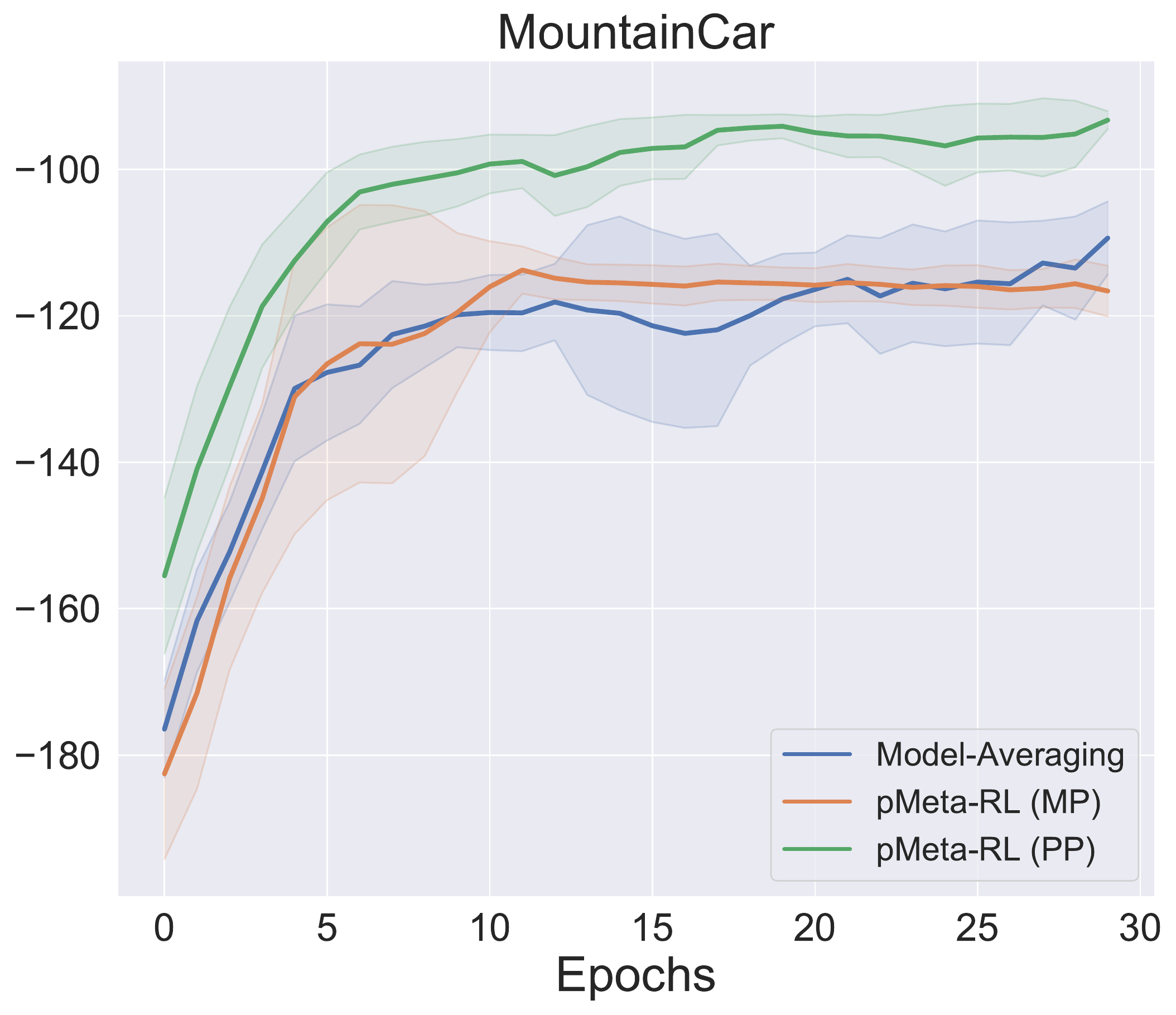}
	}
	\caption{Average Returns on the simple Gridworld and the Gym suite. (Concrete lines: the average over 5 seeds; Shaded areas: the standard deviation over 5 seeds). MP and PP represent the meta-policy and personalized policy, respectively. \textbf{Top:} Average returns under different algorithms and different $\lambda$. \textbf{Bottom:} Average returns with different algorithms on the Gym suite \cite{1606.01540}.}
	\label{fig_pFedRL}
\end{figure}
To evaluate the performance of the proposed pMeta-RL, we construct a simple Gridworld environment with a finite state-action space based on \cite{gym_minigrid}. 
In this Gridworld environment, the goal of each task is to reach the landmark in the grid and the reward is set as the distance between the agent and landmark. 
The number of tasks is $N=5$, and the tasks vary in different grid sizes. We use the model averaging method of the Q-values as the baseline, whose performance of this method has been validated in \cite{anschel2017averaged}. 
In particular, this method only performs weighted average aggregation on the Q tables of each task to obtain the multi-task Q-table. We set $C=10, R=3, K=1, \lambda=10$ in the experiments. 
Figure \ref{fig_pFedRL} shows the convergence rate of pMeta-RL, the average returns obtained by the personalized policy and meta-policy are better than that obtained by the model-averaging method. 
Note that the performance of the model-averaging method can gradually approach the personalized model as the training epoch increases. This is because the multi-task Q-table also retains all the state-action values of each task, which can adapt to different tasks. Furthermore, we directly extend the Q-learning-based approach to DQN and test the performance in the Gym environments, whose tasks vary the transition function. We can find that the performance of the personalized policy is better than the averaged-DQN, which is not well adapted to distinct tasks from Figure~\ref{fig_pFedRL}.

Figure \ref{fig_pFedRL} shows the performance of pMeta-RL in Gridworld environment with different values of $\lambda$, where we set $C=10, R=3, K=1$. 
When $\lambda=10$, pMeta-RL achieves the best performance. However, a larger $\lambda$ may degrade performance which mean this parameter need to be chosen carefully for various environments.

\subsubsection{Effects of regularization $\lambda$}
Table~\ref{table_lam} shows the performance of pMeta-RL with different values of $\lambda$, where we set $\left|D_i\right|=64, C=30, \alpha=0.001, \eta=0.001, \beta=1$.
In our experiments, we found that a proper $\lambda$ can achieve better performance, while a very larger $\lambda$ may hurt performance. 
Therefore, $\lambda$ should be carefully designed for various environments. 
We hence choose $\lambda = 20$ for ``CartPole'' and ``MountainCar'' in the remaining experiments. 

\begin{table*}[htbp] \small
\caption{Effect of regularization $\lambda$}
  \centering
  \scalebox{1.0}{
    \begin{tabular}{ccccccc}
    \toprule
    \multicolumn{7}{c}{Average Return of pMeta-RL} \\
    % result on Mnist
    \midrule
    \multirow{3}{*}{CartPole} & Change $\lambda$    
                        & $\lambda=$5      & $\lambda=$20       & $\lambda=$30       & $\lambda=$50       &  \\
      &  MP  & 185.11 $\pm$ 2.52   &  \bf{192.54} $\pm$ 1.56  & 191.51 $\pm$ 1.23  & 190.86 $\pm$ 2.09           \\
      &  PP  & 196.47 $\pm$ 0.89  & \bf{197.13} $\pm$ 0.45  & 196.53 $\pm$ 1.03  & 196.72 $\pm$ 0.68            \\
    
    \midrule
    \multirow{3}{*}{MountainCar} & Change $\lambda$    
                        & $\lambda=$5      & $\lambda=$20       & $\lambda=$30       & $\lambda=$50       &  \\
      &  GM  & -111.15 $\pm$ 4.31  & \bf{-107.08}$\pm$ 3.21  & -110.49 $\pm$ 3.89  & -108.41 $\pm$ 2.78            \\
      &  PM  & -93.05 $\pm$ 1.07   & \bf{-92.13} $\pm$ 0.95  & -92.41 $\pm$ 1.25   & -98.16 $\pm$ 2.13  \\
    \bottomrule
    \end{tabular}%
    }
  \label{table_lam}%
\end{table*}%

\begin{table}[!htbp]
  \caption{Best average return comparison results for each algorithm on the Gym suites.}
  \centering
  \scalebox{1.0}{
  \begin{tabular}{cccccccccccc}
    \toprule
    {Algorithms}  & CartPole & MountainCar   \\
    \midrule
        {DQN*}               & 198.05 $\pm$ 1.03   & -92.96 $\pm$ 2.08     \\
        {DQN}               & 169.20 $\pm$ 2.89   & -133.96 $\pm$ 3.25     \\
        {Model-Averaging}   & 191.92 $\pm$ 1.36   & -111.34 $\pm$ 5.95           \\
        {pMeta-RL (MP)}      & 192.54 $\pm$ 2.54   & -107.15 $\pm$ 4.10           \\
        {pMeta-RL (PP)}      & \bf{197.13 $\pm$ 0.45}   & \bf{-93.05  $\pm$ 0.95}         \\
    \bottomrule
  \end{tabular}}
  \label{table1}
\end{table}

\subsubsection{Ablation study}
\begin{figure*}[!tbp]
	\centering
	\subfloat{
	\includegraphics[width=7.0in]{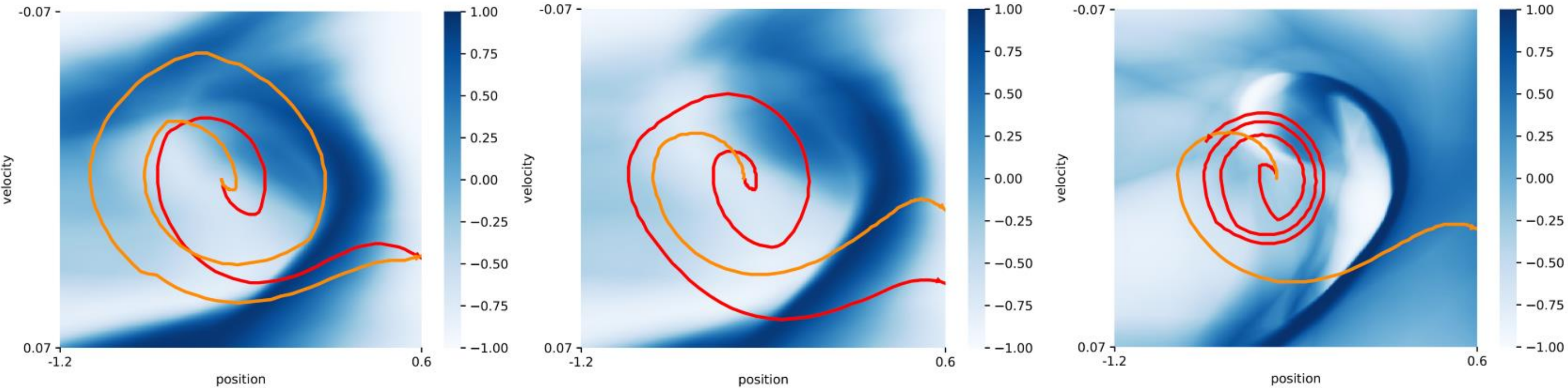}
	}
	\caption{Visualization of policies and trajectories on MountainCar. 
	The orange and red lines represent the trajectories on task 1 and task 2, respectively. The heatmaps from left to right are denoted as meta-policy, personalized policy for task 1 and optimal policy for task 1.
	\textbf{Left:} The car using the meta-policy can reach the goals (i.e., position $\geq 0.5$) on both tasks 1 and 2, but it takes more time steps on task 1. 
	\textbf{Middle:} The car using the personalized policy can reach the goals quickly on task 1, but slower on task 2.  
	\textbf{Right:} The optimal policy plans the optimal trajectory for the car on task 1, but it is unable to complete task 2.}
	\label{fig_change_visualization}
\end{figure*}

We compare the performance of the proposed personalization method with DQN \cite{mnih2015human}, Averaged-DQN \cite{anschel2017averaged}. 
Table \ref{table1} shows the best average reward achieved by various algorithms. The $\text{DQN}^{*}$ denotes the average reward obtained by the optimal DQN policy for each task, while DQN denotes the average reward achieved by the DQN policy on multiple tasks. We can find that our personalization policy is closer to the optimal policy, and our meta-policy is better than other algorithms. 
Figure \ref{fig_change_visualization} shows that the personalized policy is more suitable for its specific task than meta-policy, and illustrates the generalization ability of the personalized policy under the constraints of the meta-policy, which can generalize to other tasks while completing its corresponding task.

\subsection{Performance of Deep pMeta-RL}

\begin{figure}[!tbp]
	\centering
	\subfloat{
	\includegraphics[width=3.2in]{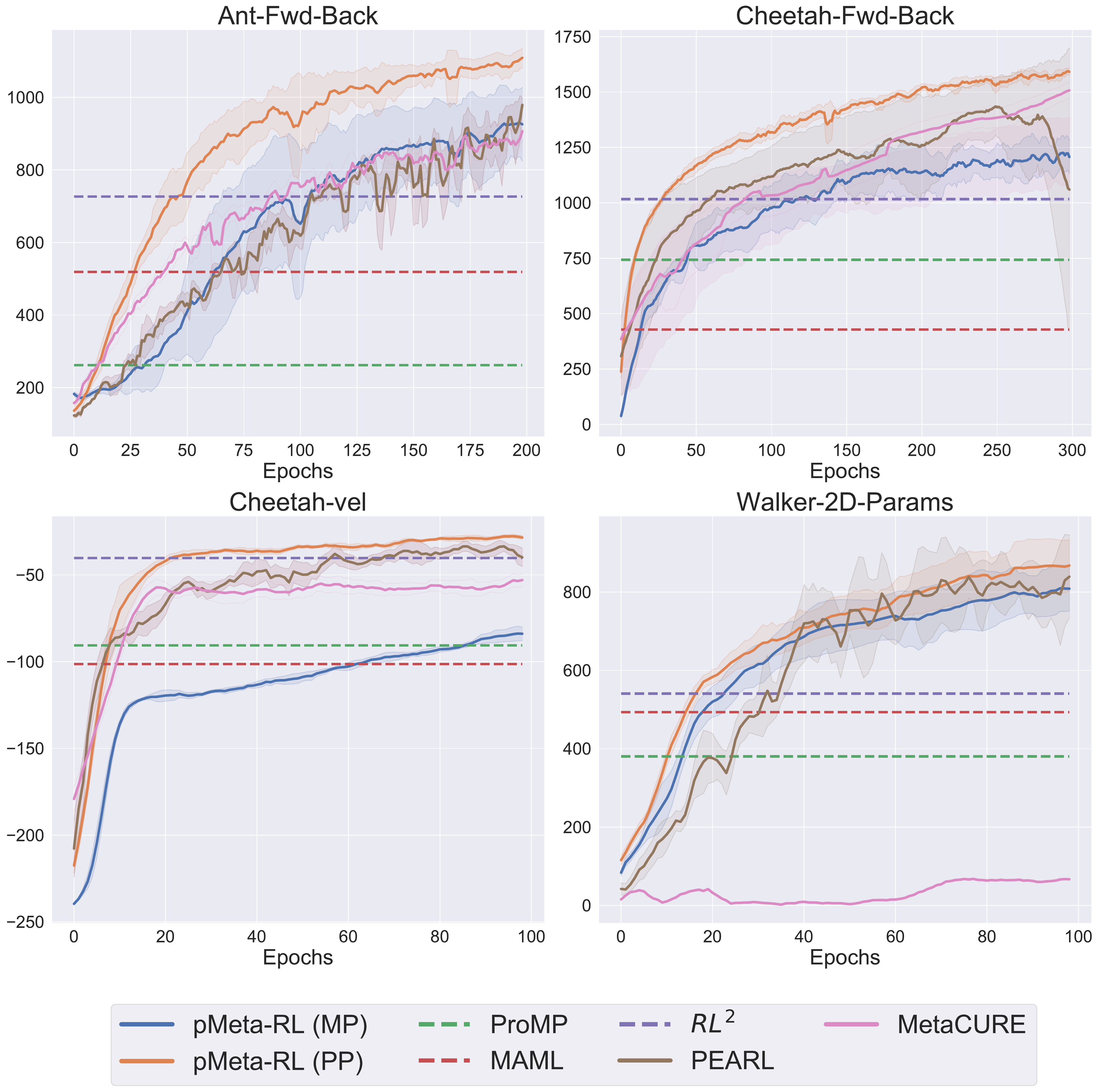}
	}
	\caption{Comparison results of the average returns of different meta-RL algorithms on the MuJoCo suites. The dotted line represents the final performance.}
	\label{fig_meta}
\end{figure}
To evaluate the proposed deep pMeta-RL algorithm, we conducted extensive experiments on the continuous control tasks in the MuJoCo suits \cite{todorov2012mujoco}, which are benchmarks commonly used by meta-RL algorithms. 
These tasks vary in either the reward function (target velocity for Half-Cheetah-Vel, and walking direction for Half-Cheetah-Fwd-Back, Ant-Fwd-Back) or transition function (Walker-Params). 
%These tasks are benchmarks commonly used by current meta-RL algorithms \cite{finn2017model, rakelly2019efficient, zhang2021metacure}. 
We compare deep pMeta-RL against several representative meta-RL algorithms, including PEARL \cite{rakelly2019efficient}, ProMP \cite{rothfuss2018promp}, MetaCURE \cite{zhang2021metacure} and MAML-TRPO \cite{finn2017model}. 
We also compare the recurrence-based policy gradient $\text{RL}^2$ method as \cite{rakelly2019efficient}. For a fair comparison, we set the maximum episode length to 200 for all the above tasks, which is similar to PEARL. In addition, we set the same number of training steps for each epoch for the above algorithms.

Figure~\ref{fig_meta} shows that the personalized policy of deep pMeta-RL based on SAC can outperform prior meta-RL methods across all domains in terms of the average returns. 
The meta-policy achieves comparable performance to other algorithms, albeit with a drop in performance in the Half-Cheetah-Vel environment.
% This may be mainly due to the increased number of tasks in this environment, resulting in slower meta-policy convergence. 
This may be mainly due to the increase in the number of tasks in this environment and the increase in the similarity between tasks, resulting in a weakened personalization ability and a slower meta-policy convergence rate.
Furthermore, we also observe that MetaCURE is unable to obtain reasonable results in Walker-2D-Params under the same setting. 
MetaCURE is an efficient algorithm for solving the sparse rewards problem. 
However, in the context of dense rewards, the intrinsic rewards proposed in the original paper may conflict with dense rewards, resulting in performance degradation. 
The average returns obtained using the personalized policies are much better than other meta-RL algorithms (e.g., $12.59\%$ and $11.86\%$ than PEARL in Ant-Fwd-Back and Cheetah-Fwd-Back environments, respectively), but only slightly gain in the Cheetah-Vel and Walker-2D-Params environments. And we found PEARL can obtains almost the same performance as the personalized policy in Walker-2D-Params. Summarizing the above, we can infer that our personalization method is better suited for more distinct tasks, e.g., navigating to the exact opposite direction.

\begin{figure}[!tbp]
	\centering
	\subfloat{
	\includegraphics[width=3.2in]{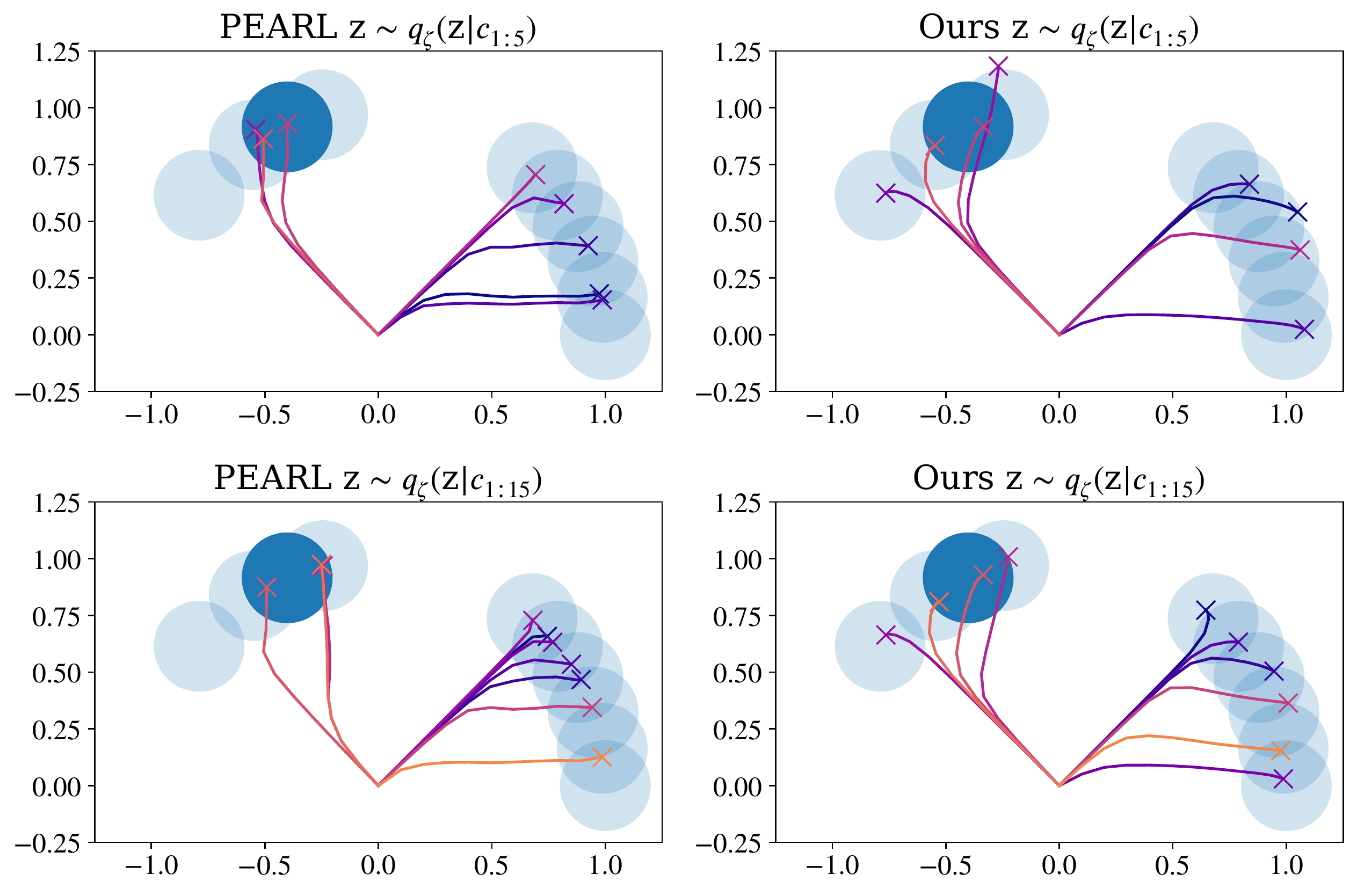}
	}
	\caption{Trajectories visualization on Sparse-Point-Robot environment. 
	The robot need to navigate to goals on a semicircle (dark blue or other goals in light blue). We compare our personalized policy with PEARL at different context $\mathbf{c}_{1:5}$ and $\mathbf{c}_{1:15}$.}
	\label{fig_vis}
\end{figure}

We visualized the trajectories to illustrate the superiority of the personalized policy on the Sparse-Point-Robot, which is a 2D navigation task in which a point robot must navigate to different goal locations on edge of a semicircle. A reward is given only when the robot is within a certain radius of the goal, which is set as 0.2 in our experiments. By randomly sampling 10 goals, we compare PEARL with different context $\mathbf{c}_{1:5}$ and $\mathbf{c}_{1:15}$, whose subscripts denote the number of trajectories contained in the context. Specifically, the first trajectory is collected with probabilistic context variable $\mathbf{z}$ sampled from the prior $p(\mathbf{z})$ and the subsequent trajectories are collected with $\mathbf{z} \sim q_{\zeta}(\mathbf{z}|\mathbf{c})$ where context is aggregated over all collected trajectories. 
As shown in Figure~\ref{fig_vis}, we can observe that that the agent with personalized policy can navigate to more targets than PEARL and achieve higher returns. Moreover, the robot can navigate more accurately with more context information, for example, a robot making decisions with $\mathbf{c}_{1:15}$ can navigate to the center of the goals, while using $\mathbf{c}_{1:5}$ may go beyond the goals radius area.

\begin{figure}[!tbp]
	\centering
	\subfloat{
	\includegraphics[width=1.5in]{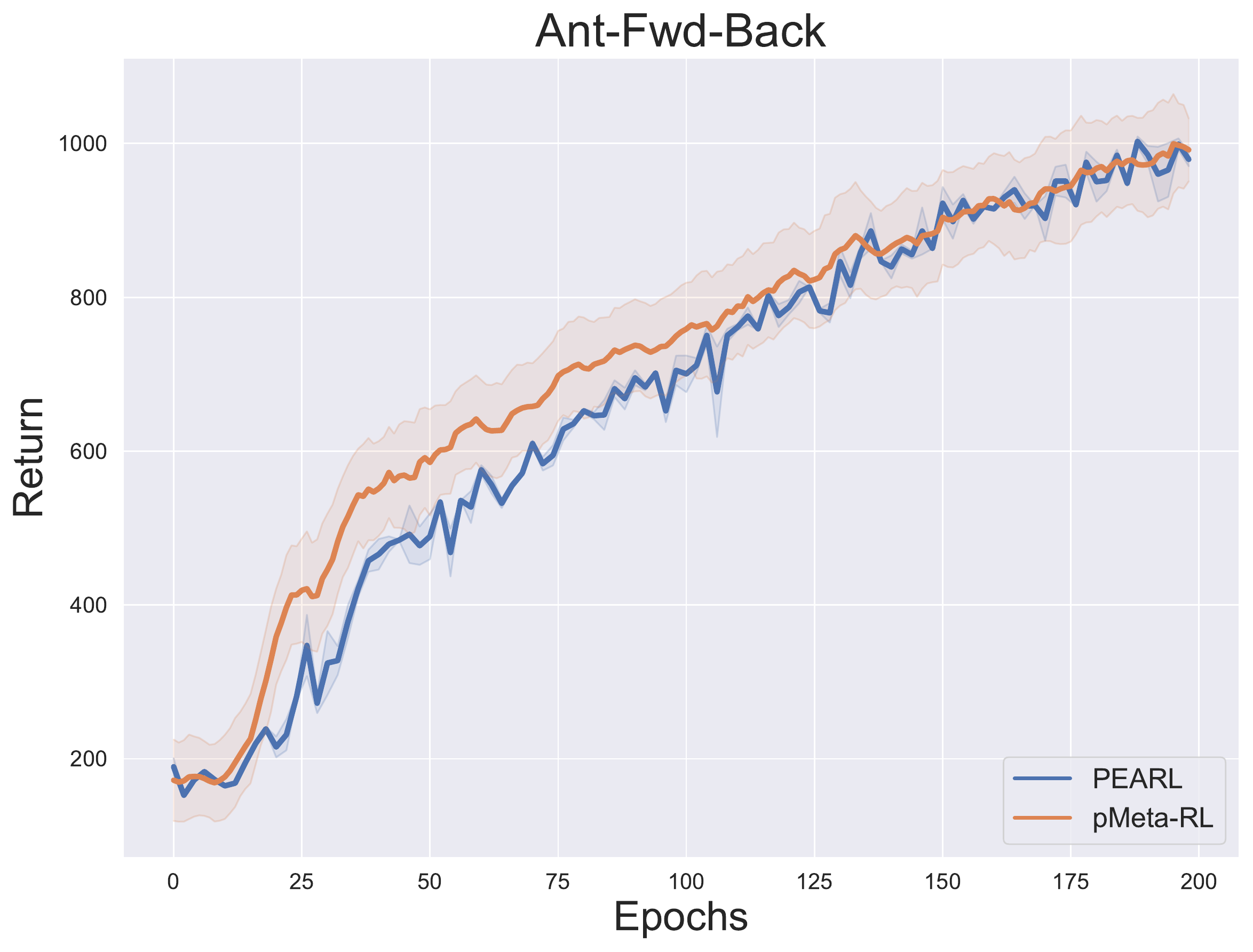}
	}
	\subfloat{
	\includegraphics[width=1.5in]{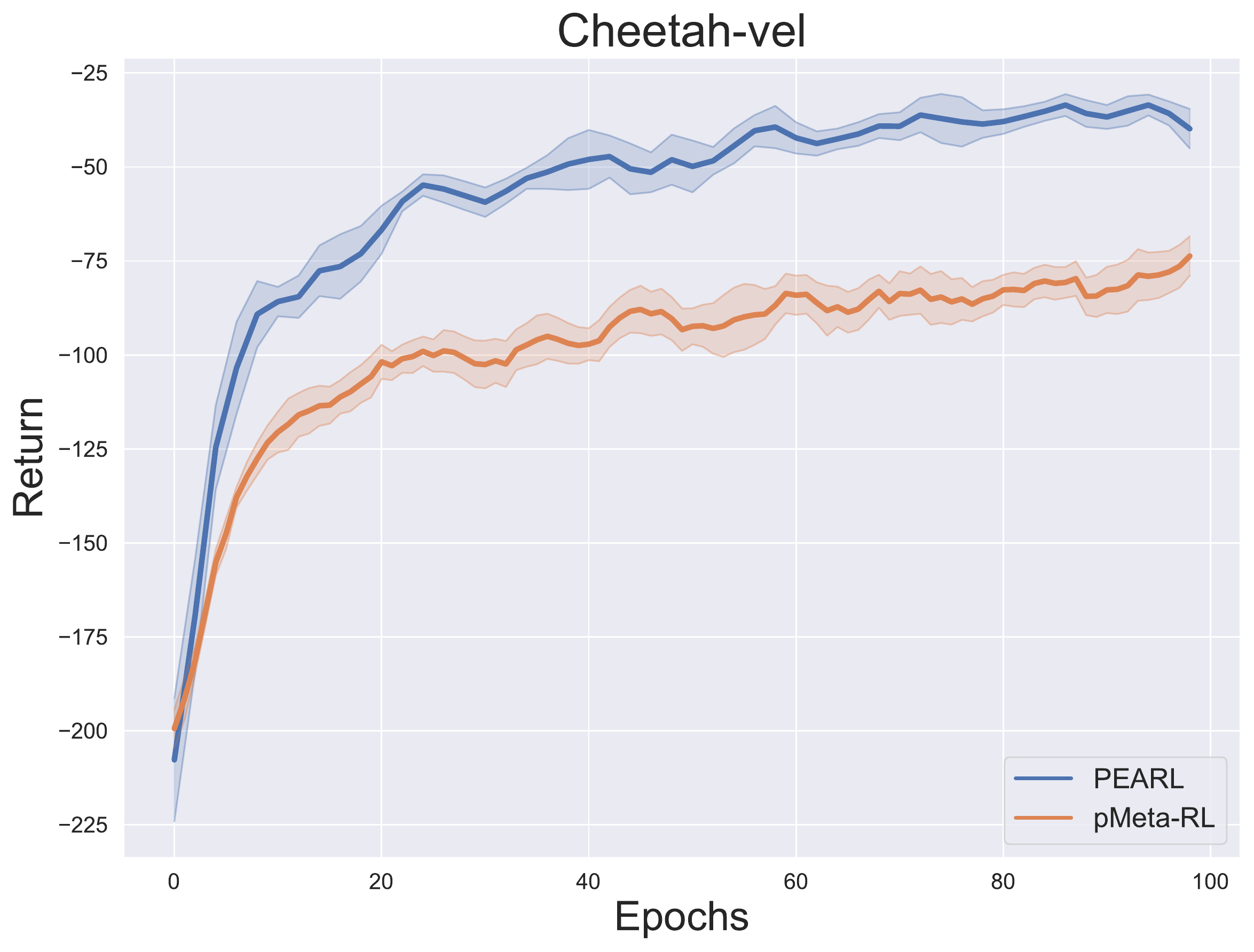}
	}
	\caption{Performance of the meta-policy on unseen tasks}
	\label{fig_test}
\end{figure}

We also validate the performance of the meta-policy on the unseen tasks in Figure~\ref{fig_test}, which is almost consistent with the performance in Figure~\ref{fig_meta}. We find that the meta-policy can adapt well to multiple tasks in most environments, except for the slower convergence on Half-Cheetah-vel. Combining Figure~\ref{fig_meta} and Figure~\ref{fig_test}, we conclude that the performance of the personalized policy is promising, such that our algorithm can be more suitable for decentralized multi-task learning, each task maintains its personalized policy while learning a meta-policy for multiple tasks without sharing trajectories.

%% file: contents/conclusion.tex
\section{Conclusions}
In this paper, we propose a personalization approach in meta-RL to solve the gradient conflict problem, which learns a meta-policy and personalized policies for all tasks and specific tasks, respectively. By adopting a personalization constrain in the objective function, our algorithm encourages each task to pursue its personalized policy around the meta-policy under the tabular and deep network settings. We introduce an auxiliary policy to decouple the personalized and meta-policy learning process and propose an alternating minimization method for policy improvement.
Moreover, theoretical analysis shows that our algorithm converges linearly with the iteration number and gives an upper bound on the difference between the personalized policies and meta-policy. Experimental results demonstrate that pMeta-RL outperforms many advanced meta-RL algorithms on the continuous control tasks.

%% file: contents/appendix.tex
\appendix
\section{Proof of Main Theorems}
\subsection{Some Useful Results}
\begin{proposition}\label{jensen}
[Jensen's inequality] For any vector $x_i \in \mathbb{R}^d, i=1,\cdots, M$, we have
\begin{equation}
    \left\|\sum_{i=1}^M x_i^2\right\| \leq M \sum_{i=1}^M\left\|x_i\right\|^2. \nonumber
\end{equation}
\end{proposition}

\begin{proposition}\label{triangle}
[Jensen's inequality] For any vector $x_1, x_2 \in \mathbb{R}^d$, we have
\begin{equation}
    \mathbb{E}\left[\left\|x_1-x_2\right\|_2^2\right] \leq \left(1+\frac{1}{R}\right)\mathbb{E}\left[\left\|x_1\right\|_2^2\right] + \left(1+R\right)\mathbb{E}\left[\left\|x_2\right\|_2^2\right], \nonumber
\end{equation}
for any constant $R$.
\end{proposition}

\begin{lemma}\label{lemma_converge}
[\cite{jaakkola1994convergence}] The random process $\left\{\Delta_t\right\}$ taking values in $\mathbb{R}^n$ and defined as \begin{equation}
    \Delta_{t+1}(x) = (1-\alpha_t(x))\Delta_t(x) + \alpha_t(x) M_t(x) \nonumber 
\end{equation}
converges to zero with probability 1 under the following assumptions:
\begin{itemize}
    \item $0 \leq \alpha_t \leq 1$, $\sum_t \alpha_t(x) =\infty$ and $\sum_t\alpha_t^2(x)<\infty$;
    \item $\left\|\mathbb{E}\left[M_t(x)|\mathcal{F}_t\right]\right\|_W \leq \gamma \left\|\Delta_t\right\|_W$, with $\gamma < 1$;
    \item $\text{\emph{var}}\left[M_t(x)|\mathcal{F}_t\right]\leq C(1+\left\|\Delta_t\right\|^2_W)$, for $C>0$.
\end{itemize}
where the $\mathcal{F}_t=\left\{\Delta_t, \Delta_{t-1},\cdots,M_{t-1},\cdots,\alpha_{t-1},\cdots\right\}$ stands for the past at step $t$, and the notation $\left\|\cdot\right\|_W$ refers to some weighted maximum norm.
    
\end{lemma}

\subsection{Some Important Lemmas}
\begin{lemma}\label{lemma_notation}
For each state-action pair, the meta Q-values update as 
\begin{equation}
\begin{split}
    Q_{c+1}^{\pi}(s,a) 
    = Q_c^{\pi}(s,a)-\tilde{\eta} Z_c(s,a)   \nonumber
\end{split}
\end{equation}
% \begin{equation}
% \begin{split}
%     Q_{c+1}^{\pi}(s,a) &= (1-\beta)Q_{c}^{\pi}(s,a) + \frac{\beta}{N}\sum_{i=1}^N Q_{i,c,R}^{\pi}(s,a) \\
%     &= Q_c^{\pi}(s,a)-\tilde{\eta} Z_c(s,a)   \nonumber
% \end{split}
% \end{equation}
where $\tilde{\eta} = \eta \beta R$ and
\begin{equation}
\begin{split}
    Z_c(s,a) = \frac{1}{NR}\sum_{i=1}^{N}\sum_{r=0}^{R-1}Z_{i,c,r}(s,a),
\end{split}
\end{equation}
with $Z_{i, c, r}(s,a)=\lambda(Q_{i,c,r}^{\pi}(s,a)-Q_{i,c,r+1}^{\pi_i}(s,a))$. 
\end{lemma}

\begin{lemma}\label{lemma_different}
Let Assumption~\ref{assumption_diversity} hold. 
For any $\tilde{Q}_i$ and $s\in\mathcal{S}, a\in \mathcal{A}$, the variance of each task is bounded 
\begin{equation}
\begin{split}
    &~~~~\frac{1}{N}\sum_{i=1}^{N}\Big\|\mathcal{R}_i(s,a)+\gamma\mathbb{E}_{s^{\prime}\sim \mathcal{P}_i(s^{\prime}|s,a)}\left[\max_{a^{\prime}}\tilde{Q}_i(s^{\prime},a^{\prime})
    \right]- \\
    & \frac{1}{N}\sum_{j=1}^N\left(\mathcal{R}_j(s,a)+\gamma\mathbb{E}_{s^{\prime}\sim \mathcal{P}_j(s^{\prime}|s,a)}\left[\max_{a^{\prime}}\tilde{Q}_i(s^{\prime},a^{\prime})
    \right]\right)\Big\|^2 \leq \sigma^2. \nonumber
\end{split}
\end{equation}
\end{lemma}

\begin{lemma}\label{lemma_diversity}
If Assumption~\ref{assumption_diversity} holds, we have
\begin{equation}
\begin{split}
   &  \frac{1}{N}\sum_{i=1}^{N}\left\|\nabla \mathcal{L}_i(Q_c^{\pi}(s,a))-\nabla \mathcal{L}(Q_c^{\pi}(s,a))\right\|^2 \\ 
   &~~~~~~~~~~~~~~~~~~~~~~~~ \leq \frac{8}{\lambda^2-8}\left\|\nabla \mathcal{L}(Q_c^{\pi}(s,a))\right\|^2+ 2 \sigma_2^2, \nonumber
\end{split}
\end{equation}
where $\sigma_2^2 = \frac{2\lambda^2}{\lambda^2-8}\sigma^2+\frac{2\lambda^2\gamma}{(\lambda^2-8)(1-\gamma)}\mathcal{R}_{\text{max}}$.
\end{lemma}

\begin{lemma}\label{lemma_drift}
The task drift error 
\begin{equation}
\begin{split}
     &~~~~~~~\frac{1}{NR}\sum_{i,r}^{N,R}\mathbb{E}\big[\left\|Z_{i,c,r}(s,a)-\nabla \mathcal{L}_{i}(Q_c^{\pi}(s,a))\right\|^2\big] \\ 
     & \leq 2\lambda^2\delta^2+ \frac{32L^2\tilde{\eta}^2}{\beta^2}\left(\frac{1}{N}\sum_{i=1}^N\mathbb{E}\big[\left\|\nabla \mathcal{L}_{i}(Q_c^{\pi}(s,a))\right\|\big]+2\lambda^2\delta^2\right). \nonumber
\end{split}
\end{equation}
\end{lemma}

\subsection{Proof of Theorem \ref{delta_converge}}\label{proof1}
We first prove that the updated form in \eqref{per_update_eq} is a contraction mapping. By setting the derivative of $\Xi_i$ to 0, we have one step optimal solution $Q_{i,c,r+1}^{\pi_i}$ as the following
\begin{equation}
\begin{split}
    &Q_{i,c,r+1}^{\pi_i}(s,a) = \frac{\lambda}{1+\lambda} Q_{i,c,r}^{\pi}(s,a) + \\ 
    &~~\frac{1}{1+\lambda} \Big(\mathcal{R}_i(s,a)+\gamma \mathbb{E}_{s^{\prime} \sim \mathcal{P}(s^{\prime}|s,a)}\left[\max_{a^{\prime}}Q_{i,c,r}^{\pi_i}(s^{\prime},a^{\prime})\right]\Big) \nonumber \\ 
\end{split}
\end{equation}
for each state-action pair $(s,a)$.

To simplify notations, we omit the subscript $c$, $r$ and $\pi_i$ from now on. 
We define an additional Bellman backup operator as
\begin{equation}\label{addition_bellman}
\begin{split}
    & \hat{\mathcal{B}}Q_i(s,a) := \frac{\lambda}{1+\lambda} Q^{\pi}(s,a) + \\ 
    & ~~\frac{1}{1+\lambda} \Big(\mathcal{R}_i(s,a) + \gamma \mathbb{E}_{s^{\prime} \sim \mathcal{P}(s^{\prime}|s,a)}\Big[ \max_{a^{\prime}}Q_i(s^{\prime},a^{\prime})\Big]\Big),
\end{split}
\end{equation}
and a norm on $Q_i$ values as $\big\|Q_i^1-Q_i^2\big\| := \max_{s,a}\big|Q_i^1(s,a)-Q_i^2(s,a)\big|$.

Suppose that $\big\|Q_i^1-Q_i^2\big\| = \varepsilon$, then we have
\begin{equation}
\begin{split}
    &~~~\mathbb{E}_{s^{\prime} \sim \mathcal{P}(s^{\prime}|s,a)}\Big[\max_{a^{\prime}}\big[Q_i^1(s^{\prime},a^{\prime})\big]\Big] \\ & \leq \mathbb{E}_{s^{\prime} \sim \mathcal{P}(s^{\prime}|s,a)}\Big[\max_{a^{\prime}}\big[Q_i^2(s^{\prime},a^{\prime})+\varepsilon\big]\Big] \\
    & \leq \mathbb{E}_{s^{\prime} \sim \mathcal{P}(s^{\prime}|s,a)}\Big[\max_{a^{\prime}}\big[Q_i^2(s^{\prime},a^{\prime})\big]\Big] + \varepsilon. \nonumber
\end{split}
\end{equation}
Therefore
\begin{equation}\label{contraction_proof}
    \left\|\hat{\mathcal{B}}Q_i^1-\hat{\mathcal{B}}Q_i^2\right\| \leq \frac{1}{1+\lambda}\gamma \varepsilon=\frac{1}{1+\lambda}\gamma \left\|Q_i^1-Q_i^2\right\|,
\end{equation}
when $Q^{\pi}(s,a)$ is constant and $\frac{\gamma}{1+\lambda}<1$. Hence $\hat{\mathcal{B}}$ is a contraction, and there exist a fixed point $Q_i^{*}(s,a)$ such that $Q_i^{*}(s,a) = \hat{\mathcal{B}}Q_i^{*}(s,a)$.

We next prove that the update method of $Q_i$ in \eqref{per_update0} can iterate to this fixed point.
We rewrite \eqref{per_update0} ignoring the superscript as the following
\begin{equation}
    \begin{split}
        &~~~~ Q_i^{k+1}(s,a) = (1-\alpha(1+\lambda))Q_i^{k}(s,a) \\
        & + \alpha\bigg[\left(\mathcal{R}_i(s,a)+\gamma \max_{a^{\prime}}Q_i^{k}(s^{\prime},a^{\prime})\right)+\lambda Q^{\pi}(s,a)\bigg] \\
        &=(1-\hat{\alpha})Q_i^{k}(s,a) \\
        & + \hat{\alpha}\bigg[\frac{1}{1+\lambda}\left(\mathcal{R}_i(s,a)+\gamma \max_{a^{\prime}}Q_i^{k}(s^{\prime},a^{\prime})\right)+\frac{\lambda}{1+\lambda} Q^{\pi}(s,a)\bigg], \nonumber
    \end{split}
\end{equation}
where $\hat{\alpha}=\alpha(1+\lambda)$ and $k$ is the iteration index.

By subtracting from both sides the fixed point $Q^{*}_i(s,a)$ and defining
\begin{equation}
    \Delta_i^k(s,a) = Q_i^k(s,a)-Q^{*}_i(s,a), \nonumber
\end{equation}
we can obtain
\begin{equation}\label{resdual_iter}
\begin{split}
    \Delta_i^{k+1}(s,a) &=(1-\hat{\alpha})\Delta_i^{k+1}(s,a) +\hat{\alpha} M_i^{k}(s,a),
\end{split}
\end{equation}
where
\begin{equation}
\begin{split}
    M_i^{k}(s,a) &= \frac{1}{1+\lambda}\left(\mathcal{R}_i(s,a) 
    + \gamma \max_{a^{\prime}}Q_i^{k}(s^{\prime},a^{\prime})\right)\\
    &~~~~~~~~~~~~~~~~~~~~~~~~+ \frac{\lambda}{1+\lambda} Q^{\pi}(s,a)-Q^{*}_i(s,a) \nonumber
\end{split}
\end{equation}
with $s^{\prime}$ being a random sample state obtained from the Markov chain $\mathcal{P}$. Then we have
\begin{equation}
\begin{split}
    & \mathbb{E}\left[M_i^k(s,a)|\mathcal{F}_k\right] = 
    \sum_{s^{\prime} \in \mathcal{S}}\mathcal{P}(s^{\prime}|s,a)\bigg[\frac{1}{1+\lambda}\bigg(\mathcal{R}_i(s,a) \\
    &~~~ +\gamma \max_{a^{\prime}}Q_i^k(s^{\prime},a^{\prime})\bigg)+\frac{\lambda}{1+\lambda} Q^{\pi}(s,a)-Q^{*}_i(s,a)\bigg]\\
    &~~~\overset{(a)}{=} \hat{\mathcal{B}}Q_i^k(s,a)-Q^{*}_i(s,a) \overset{(b)}{=} \hat{\mathcal{B}}Q_i^k(s,a)-\hat{\mathcal{B}}Q^{*}_i(s,a), \nonumber
\end{split}
\end{equation}
where $(a)$ uses the fact in \eqref{addition_bellman} and $(b)$ leverages the property of the fixed point in a contraction, i.e., $Q^{*}(s,a)=\hat{\mathcal{B}}Q^{*}(s,a)$.

According to \eqref{contraction_proof}, we have
\begin{equation}
\begin{split}
    \left\|\mathbb{E}\left[M_i^k(s,a)|\mathcal{F}_k\right]\right\|_{\infty} & \leq \frac{1}{1+\lambda}\gamma\left\|Q_i^k-Q^{*}_i\right\|_{\infty} \\ 
    &= \frac{1}{1+\lambda}\gamma\left\|\Delta_i^k\right\|_{\infty}. \nonumber
\end{split}
\end{equation}

Moreover, the variance of $M_i^k(s,a)$ can be written as 
\begin{align}\label{var}
        &~~~~~\text{var}\left[M_i^k(s,a)|\mathcal{F}_k\right] \nonumber \\
        &= \mathbb{E}\left[\left(M_i^k(s,a)-\mathbb{E}\left[M_i^k(s,a)\right]\right)^2\right] \nonumber  \\
        &= \mathbb{E}\Big[\Big(\frac{1}{1+\lambda}(\mathcal{R}_i(s,a)
        +\gamma\max_{a^{\prime}}Q_i^k(s^{\prime},a^{\prime})) \nonumber \\ 
        &~+\frac{\lambda}{1+\lambda} Q^{\pi}(s,a)-Q^{*}(s,a)-\hat{\mathcal{B}}Q_i^k(s,a)+Q^{*}(s,a)\Big)^2\Big]  \nonumber  \\
        & = \frac{1}{(1+\lambda)^2}\text{var}\Big[\Big((\mathcal{R}_i(s,a)+\gamma\max_{a^{\prime}}Q_i^k(s^{\prime},a^{\prime}))|\mathcal{F}_k\Big].
\end{align}
Since $\mathcal{R}_i(s,a)$ is bounded and $Q^{\pi}$ is constant in each personalized iteration, there exist a constant $C$ such that
\begin{equation}
    \text{var}\left[M_i^k(s,a)|\mathcal{F}_k\right]\leq \frac{1}{(1+\lambda)^2} C\left(1+\left\|\Delta^{k}\right\|^{2}\right). \nonumber
\end{equation}

To sum up, the update in \eqref{resdual_iter} satisfies all the assumptions in Lemma~\ref{lemma_converge}, therefore, $\Delta_i^k$ can converge to 0 with probability 1. In other words, there exist a $K$ such that 
\begin{equation}
    \left|Q_i^K(s,a)-Q^{*}(s,a)\right| \leq \delta. \nonumber
\end{equation}
Then we complete the proof.

\subsection{Proof of Theorem \ref{theorem_converge}} \label{proof2}
We first consider the smoothness property of $\mathcal{L}_i$. For the first term $L_i(\pi_i)=\mathbb{E}_{s \sim \rho_0}\mathbb{E}_{a \sim \pi_i}\big[Q_i^{\pi_i}(s,a)\big]$, when $\pi_i$ is Boltzmann policy \cite{sutton2018reinforcement}, i.e., $\exp{Q_i^{\pi_i}(s,a)}/\sum_a\exp{Q_i^{\pi_i}(s,a)}$, $L_i$ is a differentiable function. In addition, the regularization term $\frac{\lambda}{2} \varphi_i$ is $\ell_2$-norm, which is $\lambda$-smooth function. Thus $\mathcal{L}_i$ can be a smooth function.

We next proof the convergence of the meta-policy. Let $\mathcal{L}$ be a $L$-smoothness function, then we have 
\begin{equation}\label{proof_t1}
\begin{split}
    &~~~~~~\mathbb{E}\big[\mathcal{L}(Q_{c+1}^{\pi }(s,a))-\mathcal{L}(Q_{c}^{\pi}(s,a))\big] \\
    &\leq \mathbb{E}\big[\left \langle \nabla \mathcal{L}(Q_c^{\pi}(s,a)), (Q_{c+1}^{\pi}(s,a)-Q_c^{\pi}(s,a))\right\rangle\big] \\ &~~~~~~~~~~~~~~~~~~~~~~~~~~~~~~~+\frac{L}{2}\left\|Q_{c+1}^{\pi}(s,a)-Q_c^{\pi}(s,a)\right\|^2 \\
    &=-\tilde{\eta} \mathbb{E}\big[\left\langle \nabla \mathcal{L}(Q_c^{\pi}(s,a)), Z_c(s,a)\right\rangle\big]+\frac{\tilde{\eta}L}{2}\mathbb{E}\big[\left\|Z_c(s,a)\right\|^2\big] \\
    &=-\tilde{\eta}\mathbb{E}\big[\left\| \nabla \mathcal{L}(Q_c^{\pi}(s,a))\right\|^2\big] -\tilde{\eta} \mathbb{E}\big[\left\langle\nabla \mathcal{L}(Q_c^{\pi}(s,a)), Z_c(s,a)\right\rangle\big] \\ 
    &~~~~~~~~~~~~~~~~~~~~~~~~~~~~~~~~~~~~~~~~~~~~~+\frac{\tilde{\eta}L}{2}\mathbb{E}\big[\left\|Z_c(s,a)\right\|^2\big].
\end{split}
\end{equation}

By using Proposition \ref{jensen}, $\left\|Z_c(s,a)\right\|^2$ can be decomposed into 
\begin{equation}\label{proof_t2}
    \begin{split}
        &~~~~\mathbb{E}\big[\big\|Z_c(s,a)\big\|^2\big] \\
        & \leq 3 \mathbb{E} \Big[\Big\|\frac{1}{NR}\sum_{i,r}^{N,R}Z_{i,c,r}(s,a)-\nabla \mathcal{L}(Q_c^{\pi}(s,a))\Big\|^2\Big] \\
        &+3\Big\|\frac{1}{N}\sum_{i=1}^{N}\nabla \mathcal{L}_{i}(Q_c^{\pi}(s,a))-\nabla \mathcal{L}(Q_c^{\pi}(s,a))\Big\|^2 \\ 
        &+3\left\|\nabla \mathcal{L}(Q_c^{\pi}(s,a))\right\|^2.
    \end{split}
\end{equation}
According the Cauchy-Swartz and AM-GM inequalities, substituting \eqref{proof_t2} into \eqref{proof_t1} yields
\begin{equation}
    \begin{split}
        &~~~~\mathbb{E}\big[\mathcal{L}(Q_{c+1}^{\pi}(s,a))-\mathcal{L}(Q_c^{\pi}(s,a))\big] \\ 
        &\leq -\frac{\tilde{\eta}(1-3L\tilde{\eta})}{2}\mathbb{E}\big[\left\| \nabla \mathcal{L}(Q_c^{\pi}(s,a))\right\|^2\big]\\
        & +\frac{3\tilde{\eta}^2L}{2} \mathbb{E}\Big\|\frac{1}{N}\sum_{i=1}^N\nabla \mathcal{L}_{i}(Q_c^{\pi}(s,a))-\nabla \mathcal{L}(Q_c^{\pi}(s,a))\Big\|^2\\
        &+\frac{\tilde{\eta}(1+3\tilde{\eta} L)}{2}\frac{1}{NR}\sum_{i,r}^{N,R}\mathbb{E}\big[\left\|Z_{i,c,r}(s,a)-\nabla \mathcal{L}_{i}(Q_c^{\pi}(s,a))\right\|\big]. \nonumber
    \end{split}
\end{equation}

Then we have
\begin{align}\label{proof_con}
    &~~~~~~~\mathbb{E}\big[\mathcal{L}(Q_{c+1}^{\pi}(s,a))-\mathcal{L}(Q_c^{\pi}(s,a))\big]  \nonumber \\
    &\overset{(a)}{\leq} -\frac{\tilde{\eta}(1-3L\tilde{\eta})}{2}\mathbb{E}\big[\left\| \nabla \mathcal{L}(Q_c^{\pi}(s,a))\right\|^2\big] \nonumber  \\ 
    &~+ \frac{3\tilde{\eta}^2L}{2} \mathbb{E}\Big\|\frac{1}{N}\sum_{i=1}^N\nabla \mathcal{L}_{i}(Q_c^{\pi}(s,a))-\nabla \mathcal{L}(Q_c^{\pi}(s,a))\Big\|^2  \nonumber \\
    &~+\frac{\tilde{\eta}(1+3L\tilde{\eta})}{2}\Bigg[2\lambda^2\delta^2+\frac{32\tilde{\eta}^2 L^2}{\beta^2}\Bigg(2\lambda^2\delta^2  +\mathbb{E}\big[\left\|\nabla \mathcal{L}(Q_c^{\pi}(s,a))\right\|^2\big]\Bigg)\Bigg] \nonumber \\
    &~+\sum_{i=1}^{N}\frac{1}{N}\mathbb{E}\big[\left\|\nabla \mathcal{L}_{i}(Q_c^{\pi}{s,a})-\mathcal{L}(Q_c^{\pi}(s,a))\right\|^2\big]   \nonumber \\
    & \overset{(b)}{\leq} -\frac{\tilde{\eta}(1-3L\tilde{\eta})}{2}\mathbb{E}\big[\left\| \nabla \mathcal{L}(Q_c^{\pi}(s,a))\right\|^2\big] \nonumber \\ 
    &~ + \tilde{\eta}^2L\bigg(\frac{12}{\lambda^2-8} +\frac{16\tilde{\eta}(1+3L\tilde{\eta})\lambda^2L)}{\beta^2(\lambda^2-8)}\bigg)\mathbb{E}\left[\nabla \mathcal{L}(Q_c^{\pi}(s,a))\right]  \nonumber \\
    &~~~~~+\tilde{\eta}^3(1+3L\tilde{\eta})\frac{32L^2(\lambda^2\delta^2+\sigma^2_2)}{\beta^2}+3\tilde{\eta}^2L\sigma^2_2+\tilde{\eta}(1+3L\tilde{\eta})\lambda^2\delta^2  \nonumber \\
    & \leq -\tilde{\eta}\left[\frac{1}{2}-\tilde{\eta}L\left(\frac{3}{2}+\frac{12}{\lambda^2-8}+ \frac{24\lambda^2}{\lambda^2-8}\right)\right]\mathbb{E}\big[\left\| \nabla \mathcal{L}(Q_c^{\pi}(s,a))\right\|^2\big] \nonumber \\
    &~+\tilde{\eta}^3(1+3L\tilde{\eta})\frac{32L^2(\lambda^2\delta^2+\sigma^2_2)}{\beta^2}+3\tilde{\eta}^2L\sigma^2_2+\tilde{\eta}(1+3L\tilde{\eta})\lambda^2\delta^2  \nonumber \\
    & \overset{(c)}{\leq} -\frac{\tilde{\eta}}{2}\mathbb{E}\big[\left\| \nabla \mathcal{L}(Q_c^{\pi}(s,a))\right\|^2\big]+\frac{\tilde{\eta}^3}{\beta}96L^2(\lambda^2\delta^2+\sigma_2^2) \nonumber \\ &~+3\tilde{\eta}^2L_F\sigma^2_2+3\tilde{\eta}\lambda^2\delta^2,
\end{align}
where $(a)$ is based on Lemma~\ref{lemma_drift} and the fact that $\mathbb{E}\big[\big\|X\big\|^2\big]=\mathbb{E}\big[\big\|X-\mathbb{E}\big[X\big]\big\|^2\big]+\mathbb{E}\big[\big\|X\big\|^2\big]$ for random variable $X$, and $(b)$ is obtained by Lemma \ref{lemma_diversity}. 
When $\tilde{\eta} \leq \frac{\beta}{2L}$, we have $1+3L\tilde{\eta}\leq 1+\frac{3\beta}{2} \leq 3\beta$ and assume that
\begin{equation}
    \frac{1}{2}-\tilde{\eta}L\left(\frac{3}{2}+\frac{12}{\lambda^2-8}+ \frac{24\lambda^2}{\lambda^2-8}\right) \geq \frac{1}{4}, \nonumber
\end{equation}
then we have $(c)$.

By taking average over $C$, we have 
\begin{equation}
\begin{split}
    &~~~\frac{1}{2 C} \sum_{c=0}^{C-1} \mathbb{E}\left[\left\|\nabla \mathcal{L}\left(Q_{c}^{\pi}(s,a)\right)\right\|^{2}\right] \\ 
    &\leq \frac{\mathbb{E}\left[\mathcal{L}\left(Q_{0}^{\pi}(s,a)\right)-\mathcal{L}\left(Q^{\pi}_C(s,a)\right)\right]}{\tilde{\eta} C}+\frac{\tilde{\eta}^{2}}{\beta} Y_{4}+\tilde{\eta} Y_{5}+Y_{6}, \nonumber
\end{split}
\end{equation}
where $Y_4 = 96L^2(\lambda^2\delta^2+\sigma_2^2)$, $Y_5=3L\sigma^2_2$ and $Y_6=3\lambda^2\delta^2$.

Let $\Delta := \mathcal{L}(Q_0^{\pi}(s,a))-\mathcal{L}(Q^{g,*}(s,a))$, consider one case, when $\tilde{\eta}^3 \leq \frac{\beta \Delta}{C Y_4}$ and $\tilde{\eta}^2 \leq \frac{\Delta}{C Y_5}$, we have
\begin{equation}
\begin{split}
    & \frac{1}{2 C} \sum_{c=0}^{C-1} \mathbb{E}\left[\left\|\nabla \mathcal{L}\left(Q_{c}^{\pi}\right)\right\|^{2}\right]  \leq \frac{\Delta}{\hat{\eta}_{2} C} \\ 
    &~~~~~~~~~~~~~~~~~~+\frac{\left(\Delta\right)^{2 / 3}\left(Y_{4}\right)^{1 / 3}}{\left(\beta C\right)^{2 / 3}}+\frac{\left(\Delta Y_{5}\right)^{1 / 2}}{\sqrt{C}}+Y_{6}. \nonumber
\end{split}
\end{equation}

Finally, we proof the upper bound on the difference between the personalized polices and meta-policy,
\begin{align}\label{proof_p}
&~~~\frac{1}{N} \sum_{i=1}^{N}  \mathbb{E}\left[\left\|Q_{i,c}^{\pi_i}(s,a)-Q_{c}^{\pi}(s,a)\right\|^{2}\right] \nonumber \\
& \leq \frac{1}{N} \sum_{i=1}^{N} 2 \mathbb{E}\bigg[\left\|Q_{i,c}^{\pi_i}(s,a)-Q_{i,c}^{*}(s,a)\right\|^{2} \nonumber \\ &~~~~~~~~~~~~~~~~~~~~~~~~~~~~~+\left\|Q_{i,c}^{*}(s,a)-Q_{c}^{\pi}(s,a)\right\|^{2}\bigg] \nonumber \\
&\leq 2 \delta^{2}+\frac{2}{N} \sum_{i=1}^{N} \frac{\mathbb{E}\left[\left\|\nabla \mathcal{L}_{i}\left(Q_{c}^{\pi}(s,a)\right)\right\|^{2}\right]}{\lambda^{2}}  \nonumber \\
&\overset{(a)}{\leq} 2\delta^{2}+\frac{2}{\lambda^{2}-8} \mathbb{E}\left[\left\|\nabla \mathcal{L}\left(Q_{c}^{\pi}(s,a)\right)\right\|^{2}\right]+\frac{2 \sigma_{2}^{2}}{\lambda^{2}},
\end{align}
where $Q_{i,c}^{*}(s,a)$ is current optimal Q-values and $(a)$ is based on Lemma~\ref{lemma_diversity}.
Summing \eqref{proof_p} from $c = 0$ to $C$, we have
\begin{equation}
\begin{split}
   &~~ \frac{1}{C N} \sum_{i=0}^{C-1} \sum_{i=1}^{N} \mathbb{E}\left[\left\|Q_{i,c}^{\pi_i}(s,a)-Q_{c}^{\pi}(s,a)\right\|^{2}\right] \\ 
   & \leq \frac{2}{\lambda^{2}-8} \frac{1}{C} \sum_{i=0}^{C-1} \mathbb{E}\left[\left\|\nabla \mathcal{L}\left(Q_c^{\pi}(s,a)\right)\right\|^{2}\right]+2 \delta^{2}+\frac{2 \sigma_{2}^{2}}{\lambda^{2}}. \nonumber
\end{split}
\end{equation}
Then we complete the proof.

\subsection{Proof of Important Lemmas}
\subsubsection{Proof of Lemma \ref{lemma_notation}}
We rewrite the auxiliary Q-table $Q_i^{\pi}$ in \eqref{local_update0} update as follows 
\begin{equation}
    Q_{i,c,r+1}^{\pi}(s,a) = Q_{i,c,r}^{\pi}(s,a) - \eta \lambda(Q_{i,c,r}^{\pi}(s,a)-Q_{i,c,r+1}^{\pi_i}(s,a)), \nonumber
\end{equation}
which implies that 
\begin{equation}\label{proof_notation}
\begin{split}
    \eta \sum_{r=0}^{R-1}Z_{i,c,r}(s,a) &= \sum_{r=0}^{R-1}(Q_{i,c,r}^{\pi}(s,a)-Q_{i,c,r+1}^{\pi}(s,a))\\ 
    &=Q_{c}^{\pi}(s,a)-Q_{i,c,R}^{\pi}(s,a). 
\end{split}
\end{equation}
Note that the meta Q-values update as
\begin{equation}\label{proof_meta_Q_update}
\begin{split}
    Q_{c+1}^{\pi}(s,a) = (1-\beta)Q_{c}^{\pi}(s,a) + \frac{\beta}{N}\sum_{i=1}^N Q_{i,c,R}^{\pi}(s,a), 
\end{split}
\end{equation}
By substituting \eqref{proof_notation} into \eqref{proof_meta_Q_update}, we finish the proof.

\subsubsection{Proof of Lemma \ref{lemma_different}}
Note that $\mathcal{R}_i$ is bounded reward function such that $Q_i(s,a)$ is bounded by $\frac{\mathcal{R}_{\max}}{1-\gamma}$, then we have
\begin{equation}
\begin{split}
    &\frac{1}{N}\sum_{i=1}^{N}\Big\|\mathcal{R}_i(s,a)+\gamma\mathbb{E}_{s^{\prime}\sim \mathcal{P}_i(s^{\prime}|s,a)}\left[\max_{a^{\prime}}\tilde{Q}_i(s^{\prime},a^{\prime})
    \right] \\ 
    &~~~-\frac{1}{N}\sum_{j=1}^N\left(\mathcal{R}_j(s,a)+\gamma\mathbb{E}_{s^{\prime}\sim \mathcal{P}_j(s^{\prime}|s,a)}\left[\max_{a^{\prime}}\tilde{Q}_i(s^{\prime},a^{\prime})
    \right]\right)\Big\|^2 \\ 
    &~~~~~~~~~~~~~~~~~~~~~~~~~ \overset{(a)}{\leq} \frac{1}{N}\sum_{i=1}^{N}\Big\| \sigma_{1,i}+ \frac{\sigma_{2,i}\gamma }{1-\gamma}\mathcal{R}_{\text{max}}\Big\|^2 \leq \sigma^2, \nonumber
\end{split}
\end{equation}
where $(a)$ is based on Assumption~\ref{assumption_diversity}. We finish the proof.

\subsubsection{Proof of Lemma \ref{lemma_diversity}}
For $\mathcal{L}_i$ is a smooth function, we have $\nabla \mathcal{L}_i(Q^{\pi}(s,a)) = \lambda(Q^{\pi}(s,a)-Q_i^{*}(s,a))$. Then
\begin{align}
    &~\left\|\nabla \mathcal{L}_i(Q^{\pi}(s,a))-\nabla \mathcal{L}(Q^{\pi}(s,a))\right\|^2 \nonumber \\
    = &~ \Big\|\lambda\left(Q^{\pi}(s,a)-Q_i^{*}(s,a)\right)-\frac{1}{N}\sum_{j=1}^N\lambda\left(Q^{\pi}(s,a)-Q_j^{*}(s,a)\right)\Big\|^2 \nonumber \\
     \overset{(a)}{=}&~ \Big\|\mathcal{R}_i(s,a)+\gamma\mathbb{E}_{s^{\prime}\sim \mathcal{P}_i}\left[\max_{a^{\prime}}Q_i^{*}(s^{\prime},a^{\prime})\right]-Q_{i}^{*}(s,a)\nonumber \\
    &-\frac{1}{N}\sum_{j=1}^N\left(\mathcal{R}_j(s,a)+\gamma\mathbb{E}_{s^{\prime}\sim \mathcal{P}_j}\left[\max_{a^{\prime}}Q_j^{*}(s^{\prime},a^{\prime})\right]-Q_{j}^{*}(s,a)\right)\Big\|^2 \nonumber \\
    \overset{(b)}{\leq}&~ 2\Big\|\mathcal{R}_i(s,a)+\gamma\mathbb{E}_{s^{\prime}\sim \mathcal{P}_i}\left[\max_{a^{\prime}} Q_i^{*}(s^{\prime},a^{\prime})\right]-Q_{i}^{*}(s,a) \nonumber\\
    &-\frac{1}{N}\sum_{j=1}^N\left(\mathcal{R}_j(s,a)+\gamma\mathbb{E}_{s^{\prime}\sim \mathcal{P}_j}\left[\max_{a^{\prime}}Q_i^{*}(s^{\prime},a^{\prime})\right]-Q_{i}^{*}(s,a)\right)\Big\|^2\nonumber\\
    &+2\Big\|\frac{1}{N}\sum_{j=1}^N\left(\mathcal{R}_j(s,a)+\gamma\mathbb{E}_{s^{\prime}\sim \mathcal{P}_j}\left[\max_{a^{\prime}}Q_i^{*}(s^{\prime},a^{\prime})\right]-Q_{i}^{*}(s,a)\right)\nonumber\\
    &-\frac{1}{N}\sum_{j=1}^N\left(\mathcal{R}_j(s,a)+\gamma\mathbb{E}_{s^{\prime}\sim \mathcal{P}_j}\left[\max_{a^{\prime}}Q_j^{*}(s^{\prime},a^{\prime})\right]-Q_{j}^{*}(s,a)\right)\Big\|^2, \nonumber
\end{align}
% \begin{align}
%     &~\left\|\nabla \mathcal{L}_i(Q^{\pi}(s,a))-\nabla \mathcal{L}(Q^{\pi}(s,a))\right\|^2 \nonumber \\
%     = &~ \Big\|\lambda\left(Q^{\pi}(s,a)-Q_i^{*}(s,a)\right)-\frac{1}{N}\sum_{j=1}^N\lambda\left(Q^{\pi}(s,a)-Q_j^{*}(s,a)\right)\Big\|^2 \nonumber \\
%      \overset{(a)}{=}&~ \Big\|\mathcal{R}_i(s,a)+\gamma\mathbb{E}_{s^{\prime}\sim \mathcal{P}_i(s^{\prime}|s,a)}\left[\max_{a^{\prime}}Q_i^{*}(s^{\prime},a^{\prime})\right]-Q_{i}^{*}(s,a)\nonumber \\
%     &-\frac{1}{N}\sum_{j=1}^N\left(\mathcal{R}_j(s,a)+\gamma\mathbb{E}_{s^{\prime}\sim \mathcal{P}_j(s^{\prime}|s,a)}\left[\max_{a^{\prime}}Q_j^{*}(s^{\prime},a^{\prime})\right]-Q_{j}^{*}(s,a)\right)\Big\|^2 \nonumber \\
%     \overset{(b)}{\leq}&~ 2\Big\|\mathcal{R}_i(s,a)+\gamma\mathbb{E}_{s^{\prime}\sim \mathcal{P}_i(s^{\prime}|s,a)}\left[\max_{a^{\prime}} Q_i^{*}(s^{\prime},a^{\prime})\right]-Q_{i}^{*}(s,a) \nonumber\\
%     &-\frac{1}{N}\sum_{j=1}^N\left(\mathcal{R}_j(s,a)+\gamma\mathbb{E}_{s^{\prime}\sim \mathcal{P}_j(s^{\prime}|s,a)}\left[\max_{a^{\prime}}Q_i^{*}(s^{\prime},a^{\prime})\right]-Q_{i}^{*}(s,a)\right)\Big\|^2\nonumber\\
%     &+2\Big\|\frac{1}{N}\sum_{j=1}^N\left(\mathcal{R}_j(s,a)+\gamma\mathbb{E}_{s^{\prime}\sim \mathcal{P}_j(s^{\prime}|s,a)}\left[\max_{a^{\prime}}Q_i^{*}(s^{\prime},a^{\prime})\right]-Q_{i}^{*}(s,a)\right)\nonumber\\
%     &-\frac{1}{N}\sum_{j=1}^N\left(\mathcal{R}_j(s,a)+\gamma\mathbb{E}_{s^{\prime}\sim \mathcal{P}_j(s^{\prime}|s,a)}\left[\max_{a^{\prime}}Q_j^{*}(s^{\prime},a^{\prime})\right]-Q_{j}^{*}(s,a)\right)\Big\|^2, \nonumber
% \end{align}
where $(a)$ is due to the property of the fixed point of $\hat{\mathcal{B}}$ mentioned in \eqref{addition_bellman}, i.e.,
\begin{equation}
\begin{split}
     Q_i^{*}(s,a) &= \frac{1}{(1+\lambda)}\Big(\mathcal{R}_i(s,a) + \gamma \mathbb{E}_{s^{\prime} \sim \mathcal{P}_i(s^{\prime}|s,a)}\Big[ \max_{a^{\prime}}Q_i^{*}(s^{\prime},a^{\prime})\Big]\Big) \\ 
     &~~~~~~~~~~~~~~~~~~~~~~~~~~~~~~~~~~~~~~~`+ \frac{\lambda}{(1+\lambda)} Q^{\pi}(s,a), \nonumber
\end{split}
\end{equation}
and $(b)$ is due to Proposition \ref{jensen}. 
Taking the average over the number of tasks, we have
\begin{align}
    &~\frac{1}{N}\sum_{i=1}^N\left\|\nabla \mathcal{L}_i(Q^{\pi}(s,a))-\nabla \mathcal{L}(Q^{\pi}(s,a))\right\|^2 \nonumber \\
    \overset{(a)}{\leq} &~ 2\sigma^2 + \frac{2}{N^2}\sum_{i=1}^{N}\sum_{j=1}^N\Big\|\mathcal{R}_j(s,a) \nonumber \\
    & +\gamma\mathbb{E}_{s^{\prime}\sim \mathcal{P}_j}\left[\max_{a^{\prime}}Q_i^{*}(s^{\prime},a^{\prime})\right]-Q_{i}^{*}(s,a) \nonumber \\
    &~~-\mathcal{R}_j(s,a)+\gamma\mathbb{E}_{s^{\prime}\sim \mathcal{P}_j}\left[\max_{a^{\prime}}Q_j^{*}(s^{\prime},a^{\prime})\right]-Q_{j}^{*}(s,a)\Big\|^2 \nonumber \\
     \leq &~ 2\sigma^2 + \frac{2\gamma}{1-\gamma}\mathcal{R}_{\text{max}}+ \frac{2}{N^2}\sum_{i=1}^{N}\sum_{j=1}^N \left\|Q_{i}^{*}(s,a)-Q_{j}^{*}(s,a)\right\|^2 \nonumber \\
     \overset{(b)}{\leq} &~ 2\sigma^2 + \frac{2\gamma}{1-\gamma}\mathcal{R}_{\text{max}}+ \frac{2}{N^2}\sum_{i=1}^{N}\sum_{j=1}^N \nonumber \\ 
     &~~~~~~~~~~~~~~~\frac{2}{\lambda^2} \left(\left\|\nabla \mathcal{L}_i(Q^{\pi}(s,a))\right\|^2+\left\|\nabla \mathcal{L}_j(Q^{\pi}(s,a))\right\|^2\right) \nonumber \\
     \overset{(c)}{\leq} &~ 2\sigma^2 + \frac{2\gamma}{1-\gamma}\mathcal{R}_{\text{max}}+ \frac{8}{\lambda^2}\bigg[\left\|\nabla \mathcal{L}(Q^{\pi}(s,a))\right\|^2 \\
     &~~~~~~~~~+ \frac{1}{N}\sum_{i=1}^N\left\|\nabla \mathcal{L}_i(Q^{\pi}(s,a))-\nabla \mathcal{L}(Q^{\pi}(s,a))\right\|^2\bigg], \nonumber
\end{align}
where $(a)$ is due to Assumption~\ref{assumption_diversity} and Proposition~\ref{jensen}, and $(b)$ is based on Jensen's inequality. By re-arranging the terms in $(c)$, we obtain
\begin{equation}
\begin{split}
    & \frac{1}{N}\sum_{i=1}^N\left\|\nabla \mathcal{L}_i(Q^{\pi}(s,a))-\nabla \mathcal{L}(Q^{\pi}(s,a))\right\|^2 \\
    &~~~~~~~~~~~~~~~~~\leq \sigma_2^2 +\frac{8}{\lambda^2-8}\left\|\nabla \mathcal{L}(Q^{\pi}(s,a))\right\|^2, \nonumber
\end{split}
\end{equation}
where $\sigma_2^2 = \frac{2\lambda^2}{\lambda^2-8}\sigma^2+\frac{2\lambda^2\gamma}{(\lambda^2-8)(1-\gamma)}\mathcal{R}_{\text{max}}$.
Then we finish the proof.

\subsubsection{Proof of lemma \ref{lemma_drift}}
According to the fact $\nabla \mathcal{L}_{i}(Q_{i,c,r}^{\pi}(s,a)) = \lambda \big(Q_{i,c,r}^{\pi}(s,a)-Q_{i,c,r}^{*}(s,a)\big)$, for $r$-th iteration, we have
\begin{equation}\label{proof_drift1}
    \begin{split}
        &~~~~~~\mathbb{E}\big[\left\|Z_{i,c,r}(s,a)-\nabla \mathcal{L}_{i}(Q_{c}^{\pi}(s,a))\right\|^2\big] \\
        & \overset{(a)}{\leq} 2 \mathbb{E}\big[\left\|Z_{i,c,r}(s,a)-\nabla \mathcal{L}_{i}(Q_{i,c,r}^{\pi}(s,a))\right\|^2\big] \\
        &~~~~~~~~~~~~~+ 2\mathbb{E}\big[\nabla \mathcal{L}_{i}(Q_{i,c,r}^{\pi, g}(s,a))-\nabla \mathcal{L}_{i}(Q_{c}^{\pi}(s,a))\big] \\
        & \overset{(b)}{\leq} 2\lambda^2\mathbb{E}\big[\left\|Q_{i,c,r}^{\pi_i}(s,a)-Q_{i,c,r}^{*}(s,a)\right\|^2\big] \\ 
        &~~~~~~~~~~~~~~~~~~~~~+2 L^2\mathbb{E}\big[\left\|Q_{i,c,r}^{\pi}(s,a)-Q_c^{\pi}(s,a)\right\|^2\big]\\
        &\leq 2\Big(\lambda^2\delta^2+L^2\mathbb{E}\big[\left\|Q_{i,c,r}^{\pi}(s,a)-Q_c^{\pi}(s,a)\right\|^2\big]\Big),
    \end{split}
\end{equation}
where $(a)$ and $(b)$ are due to Proposition \ref{jensen}. 

Then we focus on the difference between $Q_{i,c,r}^{\pi}(s,a)$ and $Q_c^{\pi}(s,a)$.
The inequalities $(a), (b)$ and $(c)$ are based on the Peter Paul inequality, Proposition \ref{jensen} and \eqref{proof_drift1}, respectively. 
Let $m=R$, and when $\tilde{\eta}^2\leq\frac{\beta^2R}{4(1+R)L_F^2}$, we have $4(1+m)\eta^2 L_F^2 \leq \frac{1}{R}$ such that the inequality $(d)$ holds.
\eqref{proof_drift3} is due to unrolling \eqref{proof_drift2} recursively, and $(f)$ is according to the fact $\sum_{r=0}^{R-1}2\left(1+\frac{1}{R}\right)^r=\frac{(1+2/R)^R-1}{2/R}\leq \frac{e^2-1}{2/R}\leq4R$.

Substituting \eqref{proof_drift4} into \eqref{proof_drift1} we have
\begin{align}
    &~~~~~~\mathbb{E}\left[\left\|Q_{i,c,r}^{\pi}(s,a)-Q_c^{\pi}(s,a)\right\|\right] \nonumber \\
    &= \mathbb{E}\left[\left\|Q_{i,c,r-1}^{\pi}(s,a)-Q_c^{\pi}(s,a)-\eta  Z_{i,c,r-1}(s,a)\right\|\right] \nonumber \\ 
    &\overset{(a)}{\leq} (1+\frac{1}{m})\mathbb{E}\left\|Q_{i,c,r-1}^{\pi}(s,a)-Q_c^{\pi}(s,a)\right\|^2 \nonumber \\ 
    &~~~~~~~~~~~~~~~~~~~~~~~~~~~~~~~~~~~+(1+m)\eta^2 \mathbb{E}\left\|Z_{i,c,r-1}(s,a)\right\|^2 \nonumber \\ 
    &\overset{(b)}{\leq} (1+\frac{1}{m})\mathbb{E}\left\|Q_{i,c,r-1}^{\pi}(s,a)-Q_c^{\pi}(s,a)\right\|^2 \nonumber \\
    &~~~~~~ + 2(1+m) \eta^2\left(\mathbb{E}\left\|Z_{i,c,r-1}(s,a)-\nabla \mathcal{L}_{i}(Q_c^{\pi}(s,a))\right\|^2\right) \nonumber \\ 
    &~~~~~~~~~~~~~~~~~~~~ + 2(1+m) \eta^2 \left(\mathbb{E}\left\|\nabla \mathcal{L}_{i}(Q_c^{\pi}(s,a))\right\|^2\right) \nonumber \\ 
    &\overset{(c)}{\leq} (1+\frac{1}{m})\mathbb{E}\left\|Q_{i,c,r-1}^{\pi}(s,a)-Q_{c}^{\pi}(s,a)\right\|^2 \nonumber \\
    &~~+ 2(1+m) \eta^2\left(\mathbb{E}\left\|\nabla \mathcal{L}_{i}(Q_c^{\pi}(s,a))\right\|^2\right)  \nonumber \\ 
    &~~+4(1+m) \eta^2 
    \left(\lambda^2\delta^2+L^2\mathbb{E}\left\|Q_{i,c,r}^{\pi}(s,a)-Q_c^{\pi}(s,a)\right\|^2\right) \nonumber \\
    & = (1+\frac{1}{m}+4(1+m)\eta^2 L^2)\mathbb{E}\left\|Q_{i,c,r}^{\pi}(s,a)-Q_c^{\pi}(s,a)\right\|^2 \nonumber \\
    &~~+ 2(1+m) \eta^2 \left(\mathbb{E}\left\|\nabla \mathcal{L}_{i}(Q_c^{\pi}(s,a))\right\|^2\right)+4(1+m) \eta^2 \lambda^2\delta^2 \\  \label{proof_drift2}
    &\overset{(d)}{\leq} (1+\frac{2}{R}) \mathbb{E}\left\|Q_{i,c,r}^{\pi}(s,a)-Q_c^{\pi}(s,a)\right\|^2 \nonumber \\
    &~~+ \frac{4\tilde{\eta}^2}{\beta^2 R}\mathbb{E}\left\|\mathcal{L}_{i}(Q_c^{\pi}(s,a))\right\|^2 + \frac{8\tilde{\eta}^2\lambda^2\delta^2}{\beta^2 R} \\  \label{proof_drift3}
    &\overset{(e)}{\leq} \frac{4 \tilde{\eta}^2}{\beta^2 R}\left(\mathbb{E}\left\|\mathcal{L}_{i}(Q_c^{\pi}(s,a))\right\|^2+2\lambda^2\delta^2\right)\sum_{r=0}^{R-1}\left(1+\frac{2}{R}\right)^r  \\ \label{proof_drift4}
    &\overset{(f)}{\leq} \frac{16 \tilde{\eta}^2}{\beta^2}\left(\mathbb{E}\left\|\nabla \mathcal{L}_{i}(Q_c^{\pi}(s,a))\right\|^2+2\lambda^2\delta^2\right),   
\end{align}
then we complete the proof.